\documentclass[journal]{IEEEtran}
\ifCLASSINFOpdf
  \usepackage[pdftex]{graphicx}
  \graphicspath{{../pdf/}{../jpeg/}}
  \DeclareGraphicsExtensions{.pdf,.jpeg,.png}
\else
  \usepackage[dvips]{graphicx}
  \graphicspath{{../eps/}}
  \DeclareGraphicsExtensions{.eps}
\fi

\usepackage{amssymb}
\usepackage{amsmath}

\usepackage{color}
\usepackage{soul}

\usepackage{booktabs}
\usepackage{multirow}
\usepackage{subfigure}
\usepackage{algorithm}
\usepackage{algorithmic}

\newcommand{\INDSTATE}[1][1]{\STATE\hspace{#1\algorithmicindent}}
\usepackage{setspace}

\newcommand*{\eg}{\textit{e.g.}}
\newcommand*{\ie}{\textit{i.e.}}
\newcommand*{\etal}{\textit{et al.}}

\newcommand{\tabincell}[2]{\begin{tabular}{@{}#1@{}}#2\end{tabular}}

\begin{document}
\title{Heterogeneous Face Recognition via Face Synthesis with Identity-Attribute Disentanglement}

\author{Ziming~Yang,~
        Jian~Liang,~
        Chaoyou~Fu,~
        Mandi~Luo,\IEEEmembership{~Member,~IEEE},~
        and~Xiao-Yu~Zhang,\IEEEmembership{~Senior Member,~IEEE}~
        % <-this % stops a space
\thanks{This work was supported in part by the National Natural Science Foundation of China under Grant U2003111 and Grant 61871378, and in part by the Beijing Nova Program under Grant Z211100002121108. 
\textit{(Corresponding author: Xiao-Yu~Zhang.)}}
\thanks{Ziming Yang and Xiao-Yu Zhang are with the Institute of Information Engineering, Chinese Academy of Sciences, Beijing 100093, China, and also with the School of Cyber Security, University of Chinese Academy of Sciences, Beijing 101408, China (email:~yangziming@iie.ac.cn;~zhangxiaoyu@iie.ac.cn).}% <-this % stops a space
\thanks{Jian Liang, Chaoyou Fu, and Mandi Luo are with the National Laboratory of Pattern Recognition, Center for Research on Intelligent Perception and Computing, Institute of Automation, Chinese Academy of Sciences, Beijing 100190, China, and also with the CAS Center for Excellence in Brain Science and Intelligence Technology, Beijing 100190, China, and also with the School of Artificial Intelligence, University of Chinese Academy of Sciences, Beijing 101408, China (e-mail:~liangjian92@gmail.com;~chaoyou.fu@nlpr.ia.ac.cn;~luomandi2019@ia.ac.cn).}% <-this % stops a space
\thanks{\copyright2022 IEEE. Personal use of this material is permitted. Permission from IEEE must be obtained for all other uses, in any current or future media, including reprinting/republishing this material for advertising or promotional purposes, creating new collective works, for resale or redistribution to servers or lists, or reuse of any copyrighted component of this work in other works.}
}

\markboth{Journal of \LaTeX\ Class Files,~Vol.~14, No.~8, June~2021}%
{YANG \MakeLowercase{\textit{et al.}}: Heterogeneous Face Recognition via Face Synthesis with Identity-Attribute Disentanglement}

\maketitle

\begin{abstract}
Heterogeneous Face Recognition (HFR) aims to match faces across different domains (\eg, visible to near-infrared images), which has been widely applied in authentication and forensics scenarios. However, HFR is a challenging problem because of the large cross-domain discrepancy, limited heterogeneous data pairs, and large variation of facial attributes.
To address these challenges, we propose a new HFR method from the perspective of heterogeneous data augmentation, named Face Synthesis with Identity-Attribute Disentanglement (FSIAD).
Firstly, the identity-attribute disentanglement (IAD) decouples face images into identity-related representations and identity-unrelated representations (called attributes), and then decreases the correlation between identities and attributes.
Secondly, we devise a face synthesis module (FSM) to generate a large number of images with stochastic combinations of disentangled identities and attributes for enriching the attribute diversity of synthetic images. 
Both the original images and the synthetic ones are utilized to train the HFR network for tackling the challenges and improving the performance of HFR.
Extensive experiments on five HFR databases validate that FSIAD obtains superior performance than previous HFR approaches.
Particularly, FSIAD obtains 4.8\% improvement over state of the art in terms of VR@FAR=0.01\% on LAMP-HQ, the largest HFR database so far.
\end{abstract}


\IEEEpeerreviewmaketitle

\section{Introduction}
\IEEEPARstart{I}{n} recent years, face recognition has made significant progress with deep convolution neural networks (CNNs) and has been widely applied in real-world scenarios like surveillance and payment \cite{Masi2018, Tsai2014, Riggan2018, Kantarci2019, Huo2018,Peng2019}. Face recognition methods always assume that face images are captured with visible imaging (VIS) devices \cite{Guo2019}. However, this assumption does not hold in many realistic scenarios where face images are captured by different sensors. For instance, near-infrared (NIR) sensors are universally adopted in authentication systems and video surveillance cameras. The large discrepancy between different domains degrades face recognition performance. It arises the demand for heterogeneous face recognition (HFR) that refers to identifying faces across different domains, such as NIR-VIS, Sketch-Photo, and Thermal-VIS. Generally, HFR is confronted with three major challenges: (i) Large cross-domain discrepancy. The large discrepancy between faces in different domains enlarges intra-class distance and thus worsens the performance of HFR \cite{Pereira2019}. (ii) Lack of heterogeneous face data. It is time-consuming and expensive to collect large-scale heterogeneous face images. The limited number of subjects and small-scale HFR dataset are prone to result in the over-fitting problem \cite{He2019}. (iii) Large variation on facial attributes. Face images have various facial attributes including pose, complexion, expression, and illumination, which further increase intra-class distance and make it difficult for face matching \cite{Yu2021LAMP}. 

\begin{figure*}[ht]
\vspace{-10pt}
\centering
\includegraphics[width=0.95\linewidth]{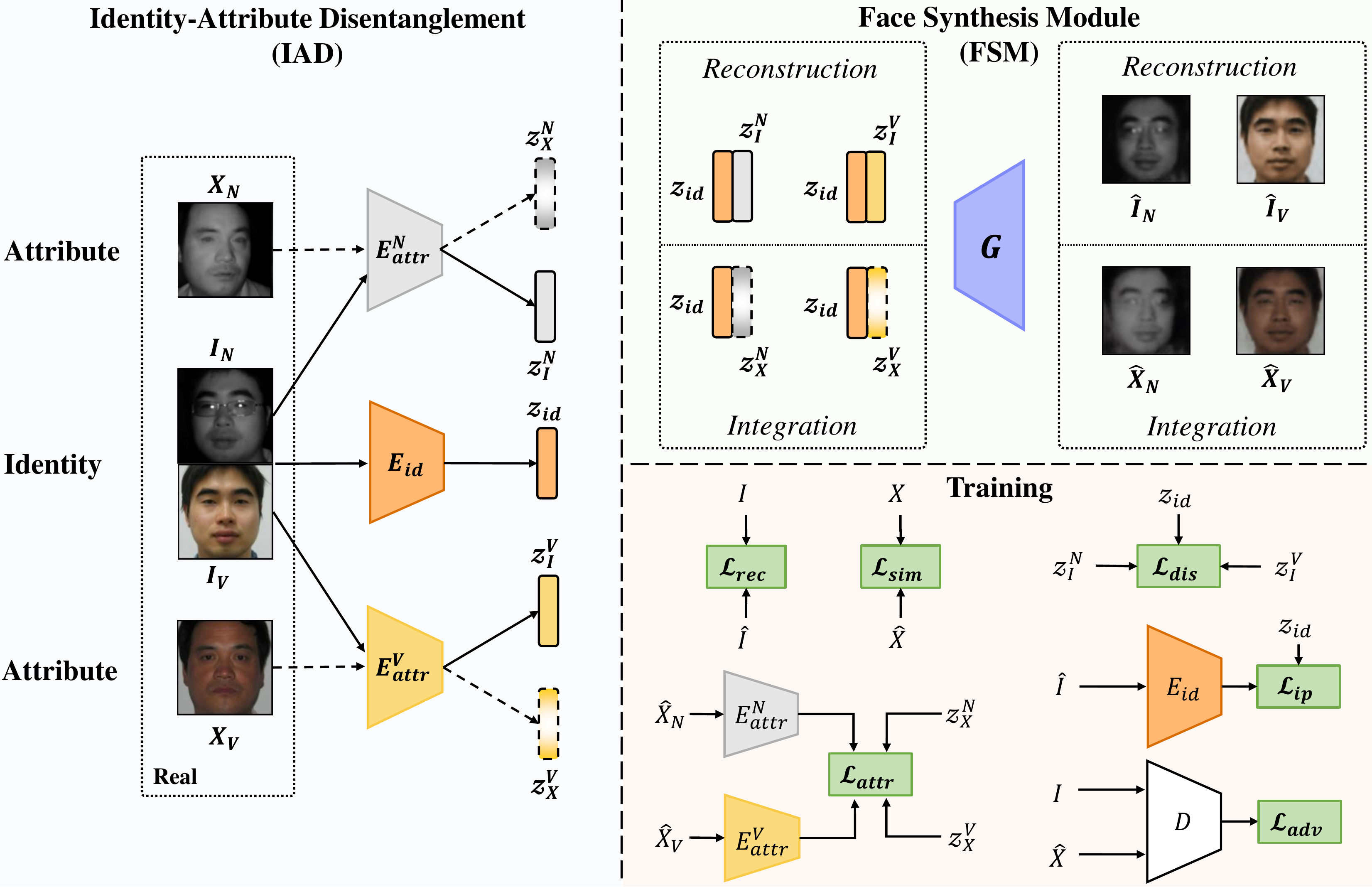}
% [width=7.1in] [width=0.95\linewidth]
% where an .eps filename suffix will be assumed under latex, 
% and a .pdf suffix will be assumed for pdflatex; or what has been declared
% via \DeclareGraphicsExtensions.
\caption{The pipeline of FSIAD. FSIAD consists of two main components, Identity-Attribute Disentanglement (IAD) and Face Synthesis Module (FSM). 
IAD: Given paired images $I=\{ I_N, I_V \}$ and $X=\{ X_N, X_V \}$, an identity encoder $E_{id}$ extracts identity representations $z_{id}$ from $I$. Meanwhile, attribute encoders $E_{attr}^{N}$ and $E_{attr}^{V}$ learn facial attribute representations from $I$ and $X$. Solid and dotted arrows indicate the processing flows of $I$ and $X$, respectively. FSM: A generator $G$ reconstructs and synthesizes images with different combinations of representations. In the training procedure, a discriminator $D$ supervises $G$ to generate high-fidelity images. The disentanglement loss $\mathcal{L}_{dis}$ makes identities uncorrelated with attributes.
The reconstruction loss $\mathcal{L}_{Rec}$ reduces the difference between reconstructed images and the corresponding input. The identity preserving loss $\mathcal{L}_{ip}$ keeps identities of $\hat{I}$ consistent with those of $I$. The attribute similarity loss $\mathcal{L}_{attr}$ and $\mathcal{L}_{sim}$ minimize the distance between attribute representations and structural features, respectively.}
\label{FSIAD}
\vspace{-10pt}
\end{figure*}

Over the last decade, HFR has attracted considerable attention of researchers to developing effective approaches \cite{ouyang2016,Jin2015}. Existing HFR methods can be divided into three main categories: domain-invariant feature based methods, common subspace learning based methods, and image synthesis based methods. The domain-invariant feature based methods seek to extract the discriminative feature of each subject that is invariant across heterogeneous domains \cite{Klare2010,Cho2021,Yang2020}. The common subspace learning based methods map heterogeneous face images into a subspace for face recognition \cite{Kan2016,Pereira2019,DADRD}. The image synthesis based methods refer to the transformation of images from one modality into the others to recognize faces within the same modality \cite{Lei2008,Lezama_2017_CVPR,Zhang2019}. Previous HFR methods mainly focus on reducing the domain discrepancy and have achieved promising performance. Nevertheless, they have neglected the variations of facial attributes in real-world applications.

To further tackle this issue, it is desirable to utilize image synthesis based methods to synthesize large numbers of images with diverse facial attributes. These synthetic images are used to train HFR models for promoting the generalization of face recognition \cite{Kortylewski2018}. Recently, \cite{Luo2021} employs data augmentation to generate images with various face deformations for alleviating the negative impacts of facial attribute variations on face recognition. Inspired by \cite{Luo2021}, we propose a novel method called Face Synthesis with Identity-Attribute Disentangle (FSIAD) to synthesize abundant heterogeneous face images with diverse facial attributes. Rather than randomly sampling from the distributions of heterogeneous data \cite{DVG}, FSIAD disentangles facial identities and attributes, and then generates a large number of images with stochastic combinations of disentangled identities and attributes to augment the raw HFR databases. As depicted in Fig. \ref{FSIAD}, the scheme of FSIAD contains two components: Identity-Attribute Disentanglement (IAD) and Face Synthesis Module (FSM). First, IAD introduces an identity encoder and attribute encoders to extract the identity-related representations and identity-unrelated representations from images, respectively. These two representations are denoted as facial identities and attributes. To disentangle attributes and identities, we regularize facial attributes to be orthogonal to identities. It decreases the correlation between facial attributes and identities and facilitates attribute representation learning. Second, FSIAD utilizes FSM to synthesize images with the combinations of disentangled identities and attributes. Massive combinations of identities and attributes significantly increase the diversity in facial attributes of synthetic images. We propose an identity preserving constraint to ensure that the identities of synthetic images are consistent with those of original images. In addition, a discriminator is applied to promote FSIAD to generate high-fidelity images by discriminating between synthetic images and real ones. The large-scale heterogeneous faces synthesized by FSIAD are used for training HFR models to fundamentally supply sufficient heterogeneous faces, learn various facial attributes and reduce the cross-domain discrepancy.

To sum up, the main contributions of this work are as follows:
\begin{itemize}
\item To improve the performance of HFR, we propose a Face Synthesis with Identity-Attribute Disentanglement (FSIAD) framework that naturally tackles three major challenges of HFR through data augmentation.
\item We devise an Identity-Attribute Disentanglement (IAD) module to decouple facial attributes and identities from face images. The facial attributes are constrained to be orthogonal to identities for decreasing their correlation.
\item We introduce a Face Synthesis Module (FSM) that generates large-scale images with the integration of disentangled facial identities and attributes to augment the HFR databases and enrich the diversity of facial attributes. 
\end{itemize}

\section{Related Work}
\label{Related Work}
Heterogeneous face recognition methods can be mainly divided into three categories, including domain-invariant feature learning, common subspace learning, and image synthesis. In this section, we review previous HFR methods and the representative generative models for image synthesis.

\subsection{Heterogeneous Face Recognition}
\subsubsection{Domain-invariant Feature Learning Methods}
\label{sec:difl}
Domain-invariant feature learning methods aim to extract the identity-related features that are invariant across spectral domains. Conventional methods are mainly supported by hand-crafted features, such as Local Binary Patterns (LBP) \cite{Ahonen2006}, Difference-of-Gaussian (DoG) \cite{Liao2009}, Histograms of Oriented Gradients (HoG) \cite{Klare2010}, and Local Binary Pattern Histogram (LBPH) \cite{Goswami2011}.

Deep learning has achieved great success in feature learning. Many efforts are devoted to extracting domain-invariant features with deep learning algorithms. The center loss \cite{Wen2016} and triple loss \cite{Liu2016} are utilized to reduce NIR-VIS discrepancy. \cite{He2019} proposes a Wasserstein CNN (W-CNN) to capture invariant deep features by minimizing the Wasserstein distance between NIR and VIS features. Based on \cite{He2019}, \cite{Wu2019} designs a Disentangled Variational Representation (DVR) framework to disentangle the identity information and within-person variations from heterogeneous faces.
\cite{Cho2021} and \cite{Cho2019} learn global relationships between local features of heterogeneous faces to represent the domain-invariant identity information.
% Hu \etal~\cite{Hu2021TMM} proposed Adversarial Disentanglement spectrum variations and Cross-modality Attention Networks (ADCANs) to take the advantage of advanced scatter loss, adversarial learning, and attention mechanism to improve the domain-invariant feature representation. 

Domain-invariant feature learning methods provide solutions to extract representations that are rarely related to facial identities. However, they are prone to suffer from the over-fitting problem due to the small-scale heterogeneous face datasets \cite{DVG-Face}. 

% Besides, these methods typically rely on handcrafted feature descriptors \cite{Klare2010, Goswami2011}, which are designed with prior knowledge on image processing and biometrics. 
\subsubsection{Common Subspace Learning Methods} Common subspace learning methods project faces of different domains into a compact latent space, where the distance between faces of the same subject is short. 
Many approaches of dimensionality reduction are used to map heterogeneous features to a low-dimensional space, including Linear Discriminant Analysis (LDA) \cite{Yi2007,Sharma2012}, Principal Component Analysis (PCA) \cite{Yi2015}, Canonical Correlation Analysis (CCA) \cite{Lei2008,Rasiwasia2010}, and Partial Least Squares (PLS) \cite{PLS}.
% Common Discriminant Feature Extraction (CDFE) \cite{Lin2006} aligned inter-modality discriminant features in a shared space and preserved local consistency for feature projection.
The emergence of deep learning has significantly promoted subspace learning.
% \cite{Kan2016} proposed a Multi-view Discriminant Analysis (MvDA) algorithm to maximize intra-view correlation and minimize inter-view correlation.
% A common encoding model \cite{Gong2017} captured discriminative features from faces in different modalities. 
Invariant Deep Representation (IDR) \cite{He2017} and Coupled Deep Learning (CDL) \cite{Wu2018} introduce orthogonal constraint and relevance constraint to learn a shared feature space, respectively.
% \cite{Pereira2019} devised Domain Specific Units to learn shallow features from faces in different domains and transform them into a generic space.
\cite{DADRD} disentangles identity-related, modality-related, and residual features from heterogeneous faces for reducing variations of modality and irrelevance (\eg pose and expression). Extended from \cite{DADRD}, \cite{OMDRA} further improves disentanglement with an orthogonal constraint and then learns residual-invariant representations by aligning high-level features of the non-neutral face and neutral face.

The common subspace learning methods intuitively project heterogeneous faces to a shared space through increasing intra-class compactness and inter-class separability. Unfortunately, since these methods require pairs of faces in different modalities for training, their performances are limited by the lack of paired heterogeneous faces. 

\subsubsection{Image Synthesis Methods} Image synthesis methods mainly include conditional and unconditional approaches. For conditional approaches, images are transformed from one modality to another modality for matching faces in the same modality. 
Face photo-sketch synthesis \cite{tang2003face,wang2009} offers the first insight into face recognition via generation.
\cite{wang2012semi,huang2013coupled,juefei2015nir} perform cross-spectral image reconstruction with coupled or joint dictionary learning.
% The advanced deep generative models especially Generative Adversarial Networks (GANs) \cite{Goodfellow2014} and Variational Anto-Encoders (VAEs) \cite{Kingma2014VAE} have made rapid progress in image synthesis. 
The advanced deep generative models \cite{Goodfellow2014,Kingma2014VAE} have made rapid progress in image synthesis. 
\cite{bae2020non} and \cite{Song2018} extend CycleGAN \cite{cyclegan} to handle heterogeneous image transformation. Later, \cite{He2020} improves \cite{Song2018} by decomposing synthesis into texture inpainting and pose correction. Pose-preserving Cross-spectral Face Hallucination (PCFH) \cite{Yu2019} and Pose Aligned Cross-spectral Hallucination (PACH) \cite{Duan2020} align the poses and expressions of NIR faces to those of VIS faces for producing paired NIR-VIS faces.
% \cite{Yu2021LAMP} introduced an exemplar-based variational spectral attention network (VSANet) to transfer the spectral features from the VIS domain to the NIR domain for image translation of heterogeneous face data.

Synthetic images produced from unconditional approaches are allowed to be inconsistent with original images.
\cite{DVG} and \cite{DVG-Face} propose a Dual Variational Generation (DVG) framework to learn a joint distribution of paired heterogeneous faces and then generate diverse heterogeneous images from noise. 
\cite{Zhu2020KD} and \cite{Hu2021TCSVT} employ knowledge distillation to use teacher networks pre-trained on abundant VIS images to train generators with insufficient paired heterogeneous faces.
Heterogeneous Face Interpretable Disentangled Representation (HFIDR) \cite{Liu2021NNLS} explicitly interprets semantic information of face representations to synthesize cross-modality face images. \cite{Di2021TBIOM} and \cite{Di2020TBIOM} synthesize multimodal faces from the predetermined visual descriptions of facial attributes.

Unlike DVR \cite{Wu2019} that only disentangles the identity information and within-person variations, we perform disentanglement and image synthesis in a unified framework. 
Contrary to DVG \cite{DVG} and DVG-Face \cite{DVG-Face} that generate the entire faces from noise and introduce external identity information from large-scale VIS faces to enrich the identity diversity of generated faces, FSIAD enrich the diversity in facial attributes of generated faces with the combinations of disentangled representations instead of external information and simultaneously tackle the challenges of HFR. 
% HFIDR \cite{Liu2021NNLS} interprets disentangled representations of identity and modality, but it can not further disentangle facial attributes from the interpreted representations. On the contrary, FSIAD disentangles features of identity, modality, and facial attributes from faces.
Compared with HFIDR \cite{Liu2021NNLS} that only interprets disentangled representations of identity and modality, FSIAD can separate representations of identity, modality, and facial attributes. 
Different from \cite{Di2021TBIOM} and \cite{Di2020TBIOM} that synthesize images from limited descriptions of facial attributes, FSIAD automatically extracts facial attributes without manual descriptions for synthesis.
Rather than decreasing variations in domains and facial attributes to learn discriminative features \cite{OMDRA}, we increase variation in facial attributes of synthetic faces to improve the generalization of HFR models to diverse attributes.

In contrast to other categories, the largest advantage of image synthesis based methods is yielding sufficient heterogeneous faces. These methods intuitively solve the lack of paired HFR data. But there are two challenges of HFR still remain: large cross-domain discrepancy and large variation of facial attributes. The performances of these methods are strongly related to the quality of synthetic images. Meanwhile, face synthesis is an ill-posed problem that there exist multiple solutions for each input \cite{Yu2021LAMP}. Our method tackles this problem by introducing comprehensive constraints to control the identities and facial attributes of synthetic faces, thereby avoiding the uncertainty of face synthesis. 

\subsection{Face Generation}
Powered by deep generative models including Generative Adversarial Networks (GANs) \cite{Goodfellow2014} and Variational Anto-Encoders (VAEs) \cite{Kingma2014VAE}, many works have achieved outstanding performance on face generation. MUNIT \cite{Huang2018MUNIT} learns content features and style features from images for multimodal image generation with unsupervised learning. Nirkin \etal~\cite{Nirkin2018} use 3D Morphable face Models (3DMM) for face segmentation and manipulation. 
\cite{Li2020} introduces a FaceShifter algorithm for face swapping and tackles the occlusion challenge to generate high-fidelity images. \cite{Karras2019,Karras2020,Karras2021} propose StyleGANs to automatically separate facial styles and stochastic effects, for improving the controllability of face synthesis. Based on \cite{Karras2020}, Nitzan \etal~\cite{Nitzan2020} propose a novel face identity disentanglement framework to disentangle identity and attribute with weakly supervised learning. Zhu \etal~\cite{Zhu2020} formulate face swapping as an optimal transfer problem, and propose an Appearance Optimal Transfer (AOT) algorithm to synthesize realistic face images with large appearance discrepancies. Daniel and Tamar \cite{Daniel2021} introduce a Soft-IntroVAE to synthesize high-resolution face images through introspective variational inference. %\cite{Huang2018}
Zhang \etal~\cite{Zhang2020} design a multi-identity face reenactment network FReeNet to transfer expressions among different subjects. InfoSwap \cite{Gao2021CVPR} utilizes the information bottleneck to extract representations of identity and perceptual features for subject-agnostic face swapping. 

\section{Method}\label{Method}
In this section, we describe the proposed method -- FSIAD in detail. As shown in Fig. \ref{FSIAD}, FSIAD contains two components: Identity-Attribute Disentanglement (IAD) and Face Synthesis Module (FSM).
First of all, IAD is designed to disentangle identity representations and facial attribute representations from face images. The identity representations are supposed to be independent of facial attribute representations. 
Secondly, FSM aims to produce large-scale heterogeneous faces with disentangled representations for improving the performance of HFR.

Our method takes two pairs of heterogeneous face images $I=\{I_N, I_V\}$ and $X=\{X_N, X_V\}$ as input data, where $N$ and $V$ denote two different spectral domains. It is applicable to NIR-VIS, Thermal-VIS, and Sketch-Photo heterogeneous faces synthesis. On the one hand, the paired images $I_{N}$ and $I_{V}$ have the same identity, which is expected to be preserved during face synthesis. On the other hand, images $\{X_N, X_V\}$ are randomly sampled from HFR datasets and provide external attribute features for enriching the diversity in attributes of generated faces.

\subsection{Identity-Attribute Disentanglement (IAD)}
The key idea of IAD is to decouple identities and attributes from faces. To extract the identity features from face images, we employ a pre-trained face recognition model LightCNN \cite{Wu2018LightCNN} as the identity encoder $E_{id}$. Latent vector $z_{id}= \frac{1}{2} (E_{id}(I_N) + E_{id}(I_V))$ contains the information merely relevant to facial identity, which is a L2-normalized feature embedding extracted by $E_{id}$. 
% Note that the identities of \{$X_N$, $X_V$\} have not been extracted since these identities are redundant for disentanglement and face synthesis. 
In addition, we design facial attribute encoders $E_{attr}^{N}$ and $E_{attr}^{V}$ based on VAEs as illustrated in Table \ref{Stc_E}. The Conv1 layer comprises a convolution layer, an instance normalization layer, and a Leaky ReLU activation layer. The FC layer denotes a fully connected layer.
The parameters of $E_{attr}^{N}$ and $E_{attr}^{V}$ are denoted as $q_{\phi_{N}}$ and $q_{\phi_{V}}$, respectively. Taking NIR image $I_N$ as input data, attribute encoder $E_{attr}^{N}$ computes the mean $\mu_I^N$ and the standard deviation $\sigma_I^N$ for approximating the posterior distribution $q_{\phi_{N}}(z_I^N|I_N) = \mathcal{N}(z_I^N; \mu_I^N, {\sigma_I^N}^2)$. The attribute representation $z_I^N$ is sampled via the re-parameterization trick, \ie,  $z_I^N = \mu_I^N + \sigma_I^N \cdot \epsilon$, where $\epsilon$ denotes the random noise that $\epsilon \sim \mathcal{N}(0,I)$. Similarly, $z_I^V$ is sampled from the posterior distribution $q_{\phi_{V}}(z_I^V|I_{V}) = \mathcal{N}(z_I^V; \mu_I^V, {\sigma_I^V}^2)$: $z_I^V = \mu_I^V + \sigma_I^V \cdot \epsilon$. 

\begin{table}[ht]
\caption{The structure of the attribute encoders $E_{attr}^{N}$ and $E_{attr}^{V}$.}
\centering
\resizebox{\linewidth}{!}{
\begin{tabular}{lcccc}
\toprule
Input & Layer & Kernel/Stride/Padding & Output & Output size \\
\midrule
image & Conv1 & 5 / 1 / 2 & $x$ & 32$\times$128$\times$128 \\

$x$ & Conv1 & 3 / 2 / 1 & $x$ & 64$\times$64$\times$64 \\

$x$ & Conv1 & 3 / 2 / 1 & $x$ & 128$\times$32$\times$32 \\

$x$ & Conv1 & 3 / 2 / 1 & $x$ & 256$\times$16$\times$16 \\

$x$ & Conv1 & 3 / 2 / 1 & $x$ & 512$\times$8$\times$8 \\

$x$ & Conv1 & 3 / 2 / 1 & $x$ & 512$\times$4$\times$4 \\

$x$ & FC & - & $\mu$, $\sigma$ & 256, 256 \\
\bottomrule
\end{tabular}
}
\label{Stc_E}
\end{table}

The training of IAD is divided into two steps, \ie, feature disentanglement and distribution learning. Feature disentanglement aims to decrease the correlation between identity and attribute representations. Hence we introduce a disentanglement objective function $\mathcal{L}_{dis}$ between identity representation $z_{id}$ and attribute representations \{$z_I^{N}$, $z_I^{V}$\}:
\begin{equation}
\mathcal{L}_{dis} = \cos({ z_{id}, z_I^{N} }) + \cos({ z_{id}, z_I^{V} }),
\end{equation}
where $\cos(\cdot,\cdot)$ is the cosine similarity function. The identity representation $z_{id}$ is orthogonal to attribute representations $z_{I}^N$ and $z_{I}^V$ when $\mathcal{L}_{dis}$ descents to 0. In other words, the learned facial attributes become independent to identity. 

The second step is designed for attribute distribution learning. Motivated by VAEs \cite{Kingma2014VAE} that are optimized through maximizing the evidence lower bound objective (ELBO), we adopt its formulation for our method as follows:
\begin{equation}\label{ELBO}
\begin{aligned}
\log{p_{\theta}}(I_N)\geq&\mathop{\mathbb{E}_{q_{\phi_{N}}(z_I^{N}|I_N)}\log{p_{\theta}}(I_N|z_I^{N})}-\\&
D_{KL}(q_{\phi_{N}}(z_I^{N}|I_N)||p(z_I^{N})),
\end{aligned}
\end{equation}
where the first term denotes the reconstruction objective of $I_N$, and the second term is a Kullback-Leibler divergence between the approximated posterior distribution $q_{\phi_{N}}(z_I^{N}|I_N)$ and the prior distribution $p(z_I^{N})$. The ELBO of $p_{\theta}(I_V)$ can be formed by simply replacing $N$ to $V$ of Eq. (\ref{ELBO}). Therefore, we define the distribution learning objective $\mathcal{L}_{kl}$ to minimize the Kullback-Leibler divergence:
\begin{equation}\label{L_kl}
\begin{aligned}
\mathcal{L}_{kl}= &D_{KL}(q_{\phi_{N}}(z_I^{N}|I_N)|| p(z_I^{N}))+\\&
D_{KL}(q_{\phi_{V}}(z_I^{V}|I_V)||p(z_I^{V})),
\end{aligned}
\end{equation}
where the prior distributions $p(z_I^{N})$ and $p({z_I^{V}})$ are assumed to obey the multivariate normal distribution $\mathcal{N}(0,I)$.

In brief, the loss function of IAD is formulated as:
\begin{equation}
\begin{aligned}
\mathcal{L}_{IAD}= \lambda_{dis}\mathcal{L}_{dis} + \mathcal{L}_{kl},
\end{aligned}
\label{L_IAD}
\end{equation}
where $\lambda_{dis}$ is a trade-off parameter. 

\subsection{Face Synthesis Module (FSM)}
As depicted in the previous section, the disentangled identities and attributes are obtained from IAD. We design an FSM to generate images with combinations of disentangled identities and attributes. 
Apart from the encoders $E_{id}$, $E_{attr}^{N}$ and $E_{attr}^{V}$ shared with IAD, FSM also contains a generator $G$ and a discriminator $D$. The structures of $G$ and $D$ are shown in TABLE \ref{Stc_G} and \ref{Stc_D}, respectively. The TransConv layer contains a transposed convolution layer, an Adaptive Instance Normalization (AdaIN) \cite{Huang2017} layer, a Leaky ReLU activation layer, and a residual block. The Tanh denotes the Tanh activation layer. The RefPad is the reflection padding operation. The Conv2 layer is comprised of a convolution layer, a batch normalization layer, and a ReLU activation layer. The Skip Connection layer adds the result $\tau$ to the input $x$. 

\begin{table}[ht]
\caption{The structure of the generator $G$.}
\centering
\resizebox{\linewidth}{!}{
\begin{tabular}{lcccc}
\toprule
Input & Layer & Kernel/Stride/Padding & Output & Output size \\
\midrule
$z_{id}$, $z_{X}$ & FC & - & $x$ & 8192 \\

$x$ & TransConv & 4 / 2 / 1 & $x$ & 256$\times$8$\times$8 \\

$x$ & TransConv & 4 / 2 / 1 & $x$ & 128$\times$16$\times$16 \\

$x$ & TransConv & 4 / 2 / 1 & $x$ & 64$\times$32$\times$32 \\

$x$ & TransConv & 4 / 2 / 1 & $x$ & 32$\times$64$\times$64 \\

$x$ & TransConv & 4 / 2 / 1 & $x$ & 32$\times$128$\times$128 \\

$x$ & Conv1 & 3 / 1 / 1 & $x$ & 32$\times$128$\times$128 \\

$x$ & Conv1 & 3 / 1 / 1 & $x$ & 3$\times$128$\times$128 \\

$x$ & Tanh & - & image & 3$\times$128$\times$128 \\
\bottomrule
\end{tabular}
}
\label{Stc_G}
\end{table}

\begin{table}[ht]
\caption{The network architecture of the Discriminator $D$.}
\label{Stc_D}
\subtable[The structure of $D$.]{
\centering
\resizebox{\linewidth}{!}{
\begin{tabular}{lcccc}
\toprule
Input & Layer & Kernel/Stride/Padding & Output & Output size \\
\midrule
image & RefPad & - & $x$ & 3$\times$134$\times$134 \\

$x$ & Conv2 & 7 / 1 / 0 & $x$ & 64$\times$128$\times$128 \\

$x$ & Conv2 & 3 / 2 / 1 & $x$ & 128$\times$64$\times$64 \\

$x$ & Conv2 & 3 / 2 / 1 & $x$ & 256$\times$32$\times$32 \\

$x$ & ResBlock & - & $x$ & 256$\times$32$\times$32 \\

$x$ & ResBlock & - & $x$ & 256$\times$32$\times$32 \\

$x$ & ResBlock & - & $x$ & 256$\times$32$\times$32 \\

$x$ & Sigmoid & - & $y$ & 1 \\
\bottomrule
\end{tabular}
}
}
\vfill
\subtable[The structure of ResBlock.]{
\centering
\resizebox{\linewidth}{!}{
\begin{tabular}{lcccc}
\toprule
Input & Layer & Kernel/Stride/Padding & Output & Output size \\
\midrule
$x$ & RefPad & - & $\tau$ & 256$\times$34$\times$34 \\

$\tau$ & Conv2 & 3 / 1 / 0 & $\tau$ & 256$\times$32$\times$32 \\

$\tau$ & RefPad & - & $\tau$ & 256$\times$34$\times$34 \\

$\tau$ & Conv2 & 3 / 1 / 0 & $\tau$ & 256$\times$32$\times$32 \\

$\tau$ & Skip Connection & - & $x$ & 256$\times$32$\times$32 \\
\bottomrule
\end{tabular}
}
}
\end{table}

The training procedure of FSM is simultaneously conducted through two branches: reconstruction and integration. The first branch aims to reconstruct images with identities and attributes that are derived from the same subjects, $\hat{I}_N=G(z_{id},z_{I}^{N})$, $\hat{I}_V=G(z_{id},z_{I}^{V})$. The second branch integrates identities of $I$ and attributes of $X$ to synthesize images $\hat{X}_{N}=G(z_{id}, z_{X}^{N})$, $\hat{X}_{V}=G(z_{id}, z_{X}^{V})$, where $z_{X}^{N}$ and $z_{X}^{V}$ are attribute representations of reference images $X_N$ and $X_V$, respectively.

\textbf{Reconstruction.} The reconstructed images \{$\hat{I}_N$, $\hat{I}_V$\} are supposed to be identical to original images \{$I_N, I_V$\}. We use the reconstruction loss $\mathcal{L}_{Rec}$ to reduce the squared Euclidean distances between original images and reconstructed ones:
\begin{equation}\label{L_rec}
\begin{aligned}
\mathcal{L}_{rec} = \Vert I_N - \hat{I}_N \Vert_2^2 + \Vert I_V - \hat{I}_V \Vert_2^2.
\end{aligned}
\end{equation}

\textbf{Integration.} In order to preserve identities of synthetic images, we extract the identity representations from synthetic images $\{\hat{I}_N, \hat{I}_V\}$ and $\{\hat{X}_N, \hat{X}_V\}$. Then we utilize identity preserving objective $\mathcal{L}_{ip}$ to increase the similarity between identity representations. $\mathcal{L}_{ip}$ is formulates as:
\begin{equation}
\begin{aligned}
\mathcal{L}_{ip} = &\Vert{z_{id} - E_{id}(\hat{I}_N)}\Vert_2^2 +
\Vert{z_{id} - E_{id}(\hat{I}_V)}\Vert_2^2 +\\&
\Vert{z_{id} - E_{id}(\hat{X}_N)}\Vert_2^2 +
\Vert{z_{id} - E_{id}(\hat{X}_V)}\Vert_2^2.
\end{aligned}
\label{L_ip}
\end{equation}

In addition, the facial attributes of synthesized images \{$\hat{X}_N$, $\hat{X}_V$\} are expected to be similar to the reference images \{$X_N$, $X_V$\}. We take both latent representations and image contents into consideration in terms of similarity. From the perspective of latent representations, the distances between attribute representations of \{$X_N$, $X_V$\} and those of \{$\hat{X}_N$, $\hat{X}_V$\} ought to be short. Similar to $\mathcal{L}_{ip}$, we define the attribute similarity objective $\mathcal{L}_{attr}$ as:
\begin{equation}
\begin{aligned}
\mathcal{L}_{attr} = \Vert{z_{X}^{N} - \hat{z}_{X}^{N}}\Vert_2^2 + \Vert{z_{X}^{V} - \hat{z}_{X}^{V}}\Vert_2^2,
\end{aligned}
\label{L_attr}
\end{equation}
where $\hat{z}_{X}^{N}$ and $\hat{z}_{X}^{V}$ are attribute representations of $\hat{X}_N$ and $\hat{X}_V$. 

Attribute-related contents including pose, complexion, and illumination, are transferred from input images into the synthesized ones. The similarity of contents can be measured from two aspects: error metric \cite{Wang2004} and structural similarity. Error metrics are prevailing approaches to evaluate the difference between images, which are calculated between images pixel by pixel. Owing to the advantage of color and illumination consistency, $L_1$-norm loss function is a predominant error metric to regularize networks for image generation. However, networks tend to generate a complete copy of the targeted image when only constrained by error metrics. Besides, error metrics are not sensitive to the pixel dependencies that contain critical structural features of objects in the visual scene \cite{Wang2003}. It results in the low perceptual quality of synthetic images. Therefore, we employ the Multi-Scale Structural Similarity index (MS-SSIM) \cite{Wang2003} as structural similarity loss to increase the structural similarity between input images and synthesized images. Specifically, MS-SSIM aims to measure the difference between structural information of images, which is related to the attributes of objects in the view and independent to extrinsic factors such as illumination, contrast, and noise \cite{Wang2004}. MS-SSIM is highly adapted for human visual perception and is generally used for image synthesis. The higher MS-SSIM means the smaller difference between objects in different images. We propose a similarity objective $\mathcal{L}_{sim}$ with an combination of a $L_1$-norm loss and MS-SSIM\footnote{https://github.com/VainF/pytorch-msssim}:
\begin{equation}
\begin{aligned}
\mathcal{L}_{sim} = (1-\alpha) \cdot \Vert{X - \hat{X}}\Vert_1 + \alpha \cdot
\mathcal{L}_{MS}(X,\hat{X}),
\end{aligned}
\label{L_sim}
\end{equation}
where $\mathcal{L}_{MS}(\cdot, \cdot)=(1-MS$-$SSIM(\cdot, \cdot))$, $\hat{X}$ means the concatenation of synthesized images, \ie, $\hat{X} =[\hat{X}_N, \hat{X}_V]$. 
$\alpha$ is a trade-off hyperparameter for tuning the ratio of $\mathcal{L}_{MS}$ to $L_1$-norm loss, which is set to 0.84 according to the empirical investigation \cite{Zhao2017}.

According to Eq. (\ref{L_ip}-\ref{L_sim}), we summarize the loss function of integration $\mathcal{L}_{Int}$ as follows:
\begin{equation}
  \begin{aligned}
  \mathcal{L}_{int} = \mathcal{L}_{ip} + \mathcal{L}_{attr} +  \mathcal{L}_{sim}.
  \end{aligned}
\label{L_Int}
\end{equation}

For the sake of further improvement on image synthesis, generator \emph{G} is optimized in an adversarial manner. We impose a discriminator \emph{D} to differentiate real images from synthetic images that generated by \emph{G}. Meanwhile, a generator \emph{G} seeks to generate photo-realistic images for confusing \emph{D}. The adversarial loss $\mathcal{L}_{adv}$ is formulated as:
\begin{equation}\label{L_adv}
\begin{aligned}
\mathcal{L}_{adv} = &\mathbb{E}_{I \sim \mathbb{P}_{data}(I)} [\log {D(I)}] +
\mathbb{E}_{\hat{X} \sim \mathbb{P}_{G}(\hat{X})} [\log({1-D(\hat{X})})]. 
\end{aligned}
\end{equation}

Hereby, we define loss function of FSM as $\mathcal{L}_{FSM}$:
\begin{equation}
\label{L_FSM}
\begin{aligned}
\mathcal{L}_{FSM} = \mathcal{L}_{rec} + \lambda_{int}\mathcal{L}_{int} + \lambda_{adv}\mathcal{L}_{adv},
\end{aligned}
\end{equation}
where $\lambda_{int}$ and $\lambda_{adv}$ are trade-off parameters.
According to Eq. (\ref{L_IAD}, \ref{L_FSM}), the overall loss of FSIAD is summarized as:
\begin{equation}\label{L_all}
\begin{aligned}
\mathcal{L}_{all} = \mathcal{L}_{IAD} + \mathcal{L}_{FSM}.
\end{aligned}
\end{equation}

\subsection{Heterogeneous Face Recognition}
We utilize FSIAD to synthesize a large number of heterogeneous faces for data augmentation. The HFR network $F$ learns abundant features of facial attributes from synthetic faces and alleviates the degradation caused by variations in pose, expression, and other extrinsic factors.
Both real images and augmented images are jointly used to train the HFR network. 

Basically, HFR network is trained on a pair of real images $\{I_N, I_V\}$. We define HFR objective $\mathcal{L}_{ce}$ with the cross-entropy loss to measure the classification error:
\begin{equation}\label{L_ce}
\begin{aligned}
\mathcal{L}_{ce} = -\sum_{i}^{n}\sum_{M\in\{N,V\}} y_i\log({\rm softmax}(F(I_M^i))),
\end{aligned}
\end{equation}
where $n$ is the number of paired real images, $y_i$ is the identity label of $i$-th paired real images $\{I_N^i, I_V^i\}$.

In addition, the HFR network is supposed to diminish the intra-class discrepancies and extract discriminative representations.
The augmented images are applied to reduce discrepancies in domains and attributes.
Since paired augmented images $\{ \tilde{X}_N, \tilde{X}_V \}$ are generated from the same subject, the identities of $\{ \tilde{X}_N, \tilde{X}_V \}$ are expected to be closed to each other \cite{DVG}. To shorten the intra-class distance, we formulate the intra-class loss $\mathcal{L}_{in}$ as:
\begin{equation}\label{L_in}
\begin{aligned}
\mathcal{L}_{in} = \Vert F(\tilde{X}_N) - F(\tilde{X}_V) \Vert_2^2.
\end{aligned}
\end{equation}
On the whole, the loss function for training HFR network is defined as:
\begin{equation}\label{L_HFR}
\begin{aligned}
\mathcal{L}_{HFR} = \mathcal{L}_{ce} + \gamma\mathcal{L}_{in},
\end{aligned}
\end{equation}
where $\gamma$ is a trade-off parameter.

For comprehensive elaboration of our work, we introduce the generic training strategy in algorithm \ref{algo}. 
\begin{algorithm}[ht]
  \renewcommand{\algorithmicrequire}{\textbf{Input:}}
  \renewcommand{\algorithmicensure}{\textbf{Output:}}
  \caption{Training strategy of FSIAD.}
  \label{algo}
  \begin{algorithmic}[1]
    \REQUIRE Source images $I$=$\{I_N, I_V\}$. Reference images $X$=$\{X_N, X_V\}$. A pre-trained identity encoder $E_{id}$. Attribute encoders $E_{attr}^N, E_{attr}^V$. A generator $G$. A discriminator $D$. A heterogeneous face recognition network $F$. The number of synthesized images $n$.
    \ENSURE The parameters of $E_{attr}^N, E_{attr}^V$, $G$, $D$ and $F$: $\phi_N$, $\phi_V$, $\Psi_G$, $\Psi_D$ and $\Theta$.
    \FOR{$i=1$ to \emph{T}}
      \STATE \texttt{// IAD component.}
      \STATE $z_{id}=\frac{1}{2}(E_{id}(I_N)+E_{id}(I_V))$.
      \STATE $z_I^N=E_{attr}^N(I_N)$; $z_I^V=E_{attr}^V(I_V)$.
      \STATE $z_X^N=E_{attr}^N(X_N)$; $z_x^V=E_{attr}^V(X_V)$.
      \STATE Update $\phi_N$, $\phi_V$ by Eq. (\ref{L_IAD}).
      \STATE \texttt{// FSM component.}
      \STATE Reconstruction: $\hat{I}_N=G(z_{id}, z_I^N)$; $\hat{I}_V=G(z_{id}, z_I^V)$.
      \STATE Integration: $\hat{X}_N=G(z_{id}, z_X^N)$; $\hat{X}_V=G(z_{id}, z_X^V)$.
      \STATE Fix $\phi_N$, $\phi_V$, $\Psi_G$:
      \INDSTATE Update $\Psi_D$ by Eq. (\ref{L_adv}).
      \STATE Fix $\Psi_D$:
      \INDSTATE Update $\phi_N$, $\phi_V$, $\Psi_G$ by Eq. (\ref{L_Int}).
    \ENDFOR 
    \STATE \texttt{// HFR network.}
    \STATE Initialize $\Theta$ by a pre-trained model.
    \FOR{$iteration=1$ to $n$}
      \STATE Randomly sample $I$=$\{I_N,I_V\}$ and $X$=$\{X_N,X_V\}$.
      \STATE $z_{id}=\frac{1}{2}(E_{id}(I_N)+E_{id}(I_V))$.
      \STATE $z_X^N$=$E_{attr}^N(X_N;\phi_N)$;   $z_X^V$=$E_{attr}^V(X_V;\phi_V)$.
      \STATE $\tilde{X}_N=G(z_{id},z_X^N;\Psi_G)$; $\tilde{X}_V=G(z_{id},z_X^V;\Psi_G)$.
      \STATE Update $\Theta$ by Eq. (\ref{L_HFR}).
    \ENDFOR
    \RETURN $\Theta$, $\phi_N$, $\phi_V$, $\Psi_G$, $\Psi_D$.
  \end{algorithmic}
\end{algorithm}

\section{Experiments}
\label{Experiments}
In this section, we perform extensive experiments to evaluate our proposed FSIAD qualitatively and quantitatively on five heterogeneous face recognition databases, including CASIA NIR-VIS 2.0 \cite{Li2013CASIA}, BUAA-VisNir \cite{Huang2012BUAA}, Oulu-CASIA NIR-VIS \cite{Chen2009Oulu}, Tufts Face \cite{Panetta2020TUFTS}, and LAMP-HQ \cite{Yu2021LAMP}. To demonstrate the superior advantage of our proposed FSIAD, we make comparisons with state-of-the-art methods. Finally, ablation studies on the effectiveness of different loss functions are conducted.

\subsection{Databases and Protocols}
\label{sec:database}
\subsubsection{CASIA NIR-VIS 2.0} the most challenging public HFR database. It contains 725 subjects and each subject has 5$\sim$50 NIR and 1$\sim$22 VIS images with large within-class variations including pose, expression, and illumination. We conduct experiments with 10-fold cross-validation. For each fold, the training set consists of about 6,100 NIR and 2,500 VIS images from 360 identities. For evaluation, the testing set is composed of over 6,000 NIR and 358 VIS images from 358 identities who are excluded from those in the training set. 

\subsubsection{BUAA-VisNir} a standard HFR database. It contains 150 subjects and each subject has 9 pairs of NIR-VIS images, including one frontal view, four different other views, and four expressions. Since the NIR and VIS images are captured simultaneously with a multi-spectral sensor, paired NIR-VIS images are identical except for the spectral domain. The training set consists of 900 images from 50 subjects and the testing set consists of 1,800 images from the remaining 100 subjects. 

\subsubsection{Oulu-CASIA NIR-VIS} contains 80 subjects. The paired NIR-VIS images of each subject are captured with six different expressions (anger, disgust, fear, happiness, sadness, and surprise) and three different illuminations (normal indoor, weak, and dark). Following the protocol introduced in \cite{Shao2017}, we select 40 subjects and each subject has 48 paired NIR-VIS images as the training set and testing set. Both the training set and the testing set comprise 20 subjects.

\subsubsection{Tufts Face} a Thermal-VIS HFR database. It contains 1,582 paired Thermal-VIS images of 113 subjects. Each subject has 14 pairs of Thermal-VIS images with 9 different yaw angles and 5 different expressions. Since this database has not designed a protocol for evaluation, we divide it into a training set with 85 subjects and a testing set with the rest 28 subjects. The training set is composed of 1,190 pairs of Thermal-VIS images, while the testing set has 28 VIS gallery images and 275 thermal probe images.

\subsubsection{LAMP-HQ} the latest and largest HFR database. It is featured with large-scale, high-resolution, and wide-diversity (\eg, age, race, and accessories), which contains 56,788 NIR and 16,828 VIS images from 573 subjects. Each subject has 66 paired NIR-VIS images that are captured with three distinct expressions and three yaw angles in five illumination scenes. The evaluation is conducted by 10-fold experiments. For each fold, the training dataset consists of approximately 29,000 NIR and 8,800 VIS images from about 300 subjects. For testing, the gallery set contains the remaining 273 individuals and each individual has one VIS image, while the probe set has about 27,000 NIR images from the same individuals.

\subsection{Experimental Settings}

\textbf{Implementation details}. We firstly transform the size of input images to $128 \times 128$. A face recognition model LightCNN \cite{Wu2018LightCNN} pre-trained on the MS-Celeb-1M dataset \cite{Guo2016} is adopted as the identity encoder $E_{id}$ to extract identity features from face images. The $L_2$ normalized latent vector output from the penultimate fully connected layer in LightCNN is utilized as identity feature representation. The attribute encoders $E_{attr_N}$, $E_{attr_V}$, generator (or decoder) $G$, and discriminator $D$ are built with the architectures that are described in TABLE \ref{Stc_E}, \ref{Stc_G} and \ref{Stc_D}, respectively. Our proposed network is implemented with the deep learning framework Pytorch. One Nvidia RTX GPU is used for acceleration. For optimization on training, we employ an Adam \cite{Adam} optimizer that is configured with a learning rate of 2e-4, coefficients $\beta_1$=0.5 and $\beta_2$=0.99 that are used to compute the running averages of gradient and its square.
The trade-off parameters $\lambda_{dis}$, $\lambda_{int}$, $\lambda_{adv}$, and $\gamma$ in Eq.(\ref{L_IAD},\ref{L_FSM},\ref{L_HFR}) are set to 2, 5, 1, and 0.001, respectively.

For fair comparisons, we follow the protocols of \cite{Liu2021NNLS,DVG,Wu2019,Duan2020,He2019} to use LightCNN \cite{Wu2018LightCNN} as the HFR model. The pre-trained LightCNN model is imported to initialize the parameters of the network. In the experiments, we utilize FSIAD to synthesize 100,000 pairs of heterogeneous faces as augmented data. Then we fine-tune the LightCNN network with the combination of the original HFR dataset and augmented data. For optimization of fine-tuning, we use the Stochastic Gradient Descent algorithm (SGD) with the following configurations: momentum factor=0.9, initial learning rate=1e-3, and weight decay=1e-4. To tune the hyper-parameters of our proposed loss functions, we split the training data of each HFR database into a training set and a validation set by a ratio of 9:1.

\subsection{Qualitative Analyses}
\begin{figure}[ht]
\centering
\includegraphics[width=\linewidth]{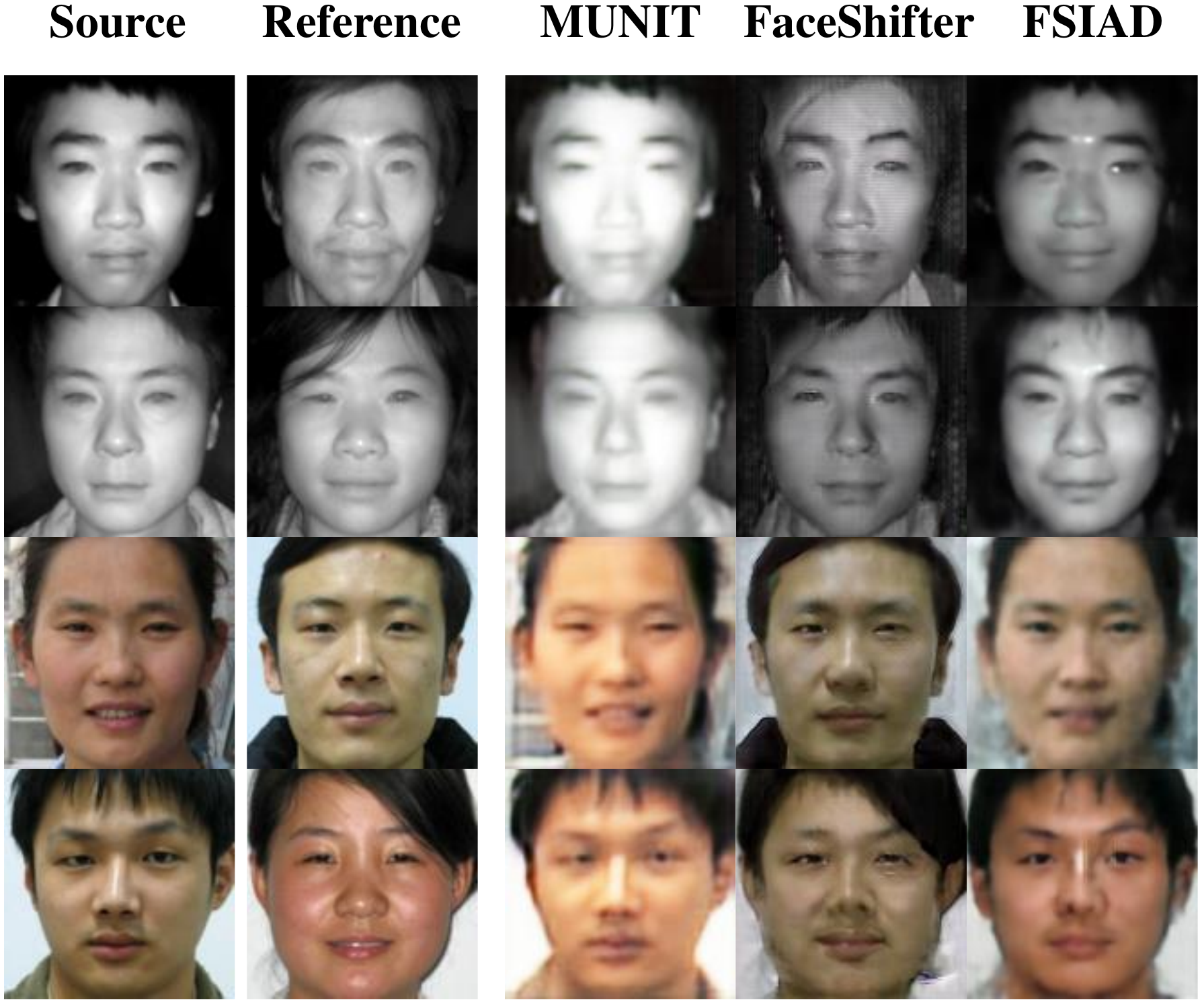}
\caption{Experimental results on feature disentanglement and generation with FSIAD and state-of-the-art methods MUNIT and FaceShifter.}
\label{Compare}
\end{figure}

In order to analyze the effectiveness of face augmentation, we conduct qualitative experiments on the CASIA NIR-VIS 2.0 database \cite{Li2013CASIA} to evaluate our proposed FSIAD with the comparison of state-of-the-art methods MUNIT \cite{Huang2018MUNIT} and FaceShifter \cite{Li2020}. Both the disentanglement and generation of face augmentation are evaluated in the qualitative experiments. We firstly decouple face images into identity representations and attribute representations. Then we produce images of $128 \times 128$ resolution with stochastic combinations of identities and attributes. The results generated by the aforementioned methods are displayed in Fig. \ref{Compare}. Intuitively, although MUNIT produces clear results, it fails to disentangle identities from attributes. The synthetic images of MUNIT are almost equal to the source images and have low similarity in facial attributes compared with the reference images. Additionally, we can observe that FaceShifter can swap identity features from the subject of the source image to that of the reference image. However, the results of FaceShifter are mixed with artifacts and lack the smoothness of facial textures. Although FaceShifter \cite{Li2020} has gained outstanding performance of face generation on the large-scale VIS databases, the insufficient HFR data hinders it from generating high-quality heterogeneous faces. In contrast to these methods, FSIAD achieves better performance in feature disentanglement and generation. For one thing, the identities of source images are disentangled from other facial attributes, which are well preserved in the synthetic images. For another, our method can well blend the identities with facial attributes and prevent the emergence of artifacts and rough textures. Moreover, facial attributes such as expression, pose, and complexion are transferred from reference images to the generated images. 

% Compare

\begin{figure}[ht]
\centering
\includegraphics[width=\linewidth]{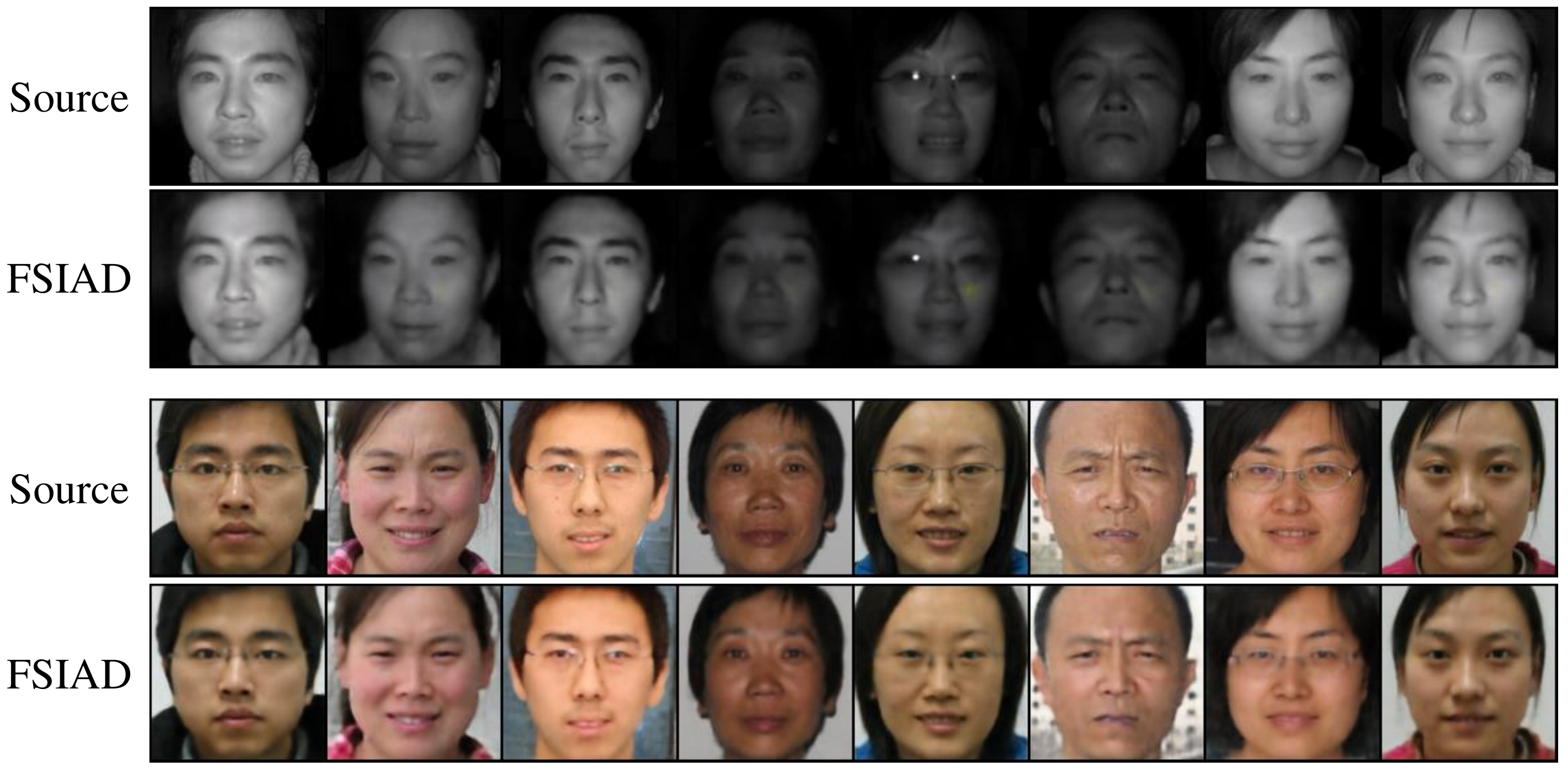}
\caption{An example of face reconstruction on the CASIA NIR-VIS 2.0 database.}
\label{Reconstruct}
\end{figure}

\begin{figure}[ht]
\centering
\includegraphics[width=\linewidth]{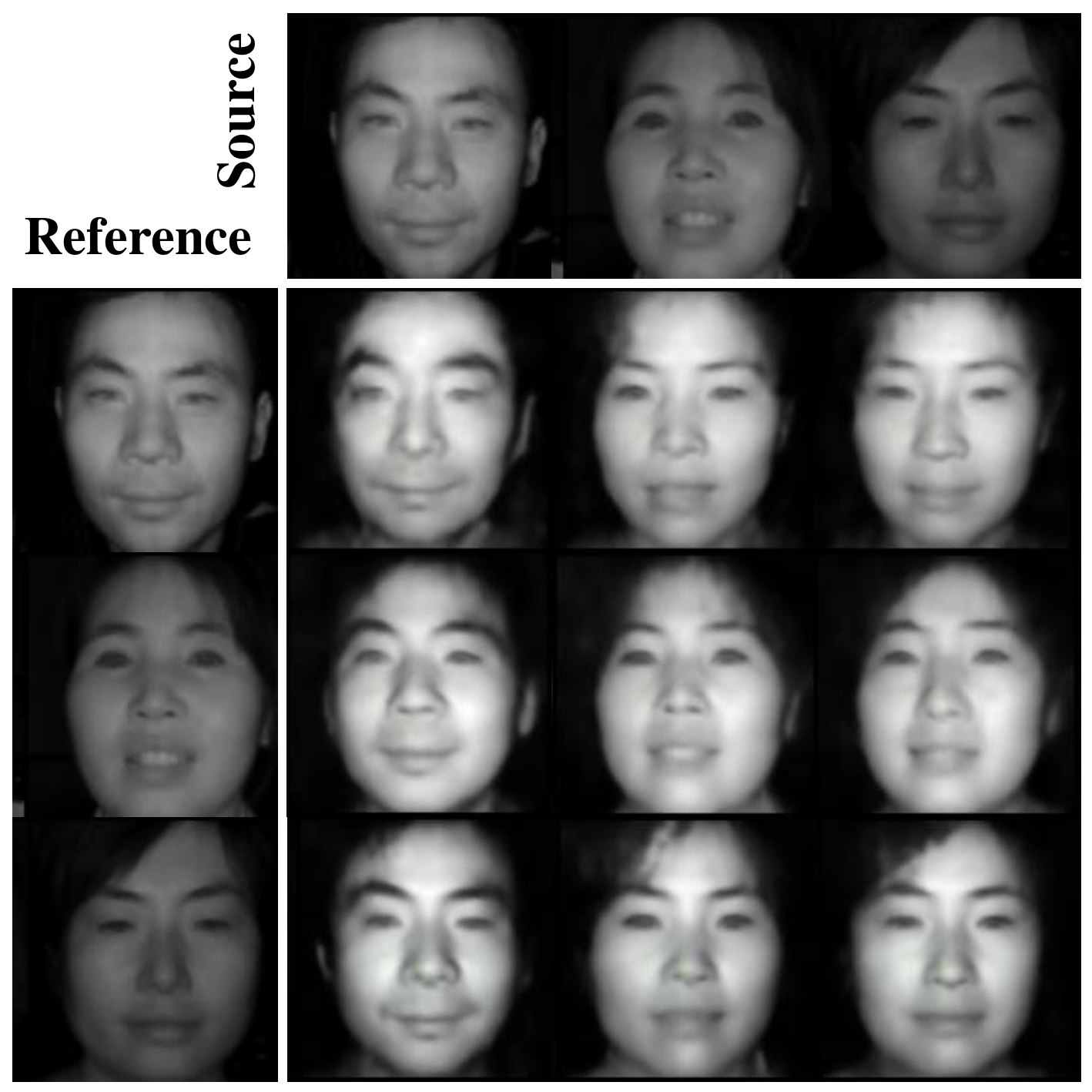}
\caption{A grid of synthetic faces. The images in the top row and left most column are source images and reference images, respectively. The rest images are synthesized by FSIAD with identities of source images and attributes of reference images.}
\label{Cross}
\end{figure}

\begin{figure}[ht]
\centering
\includegraphics[width=\linewidth]{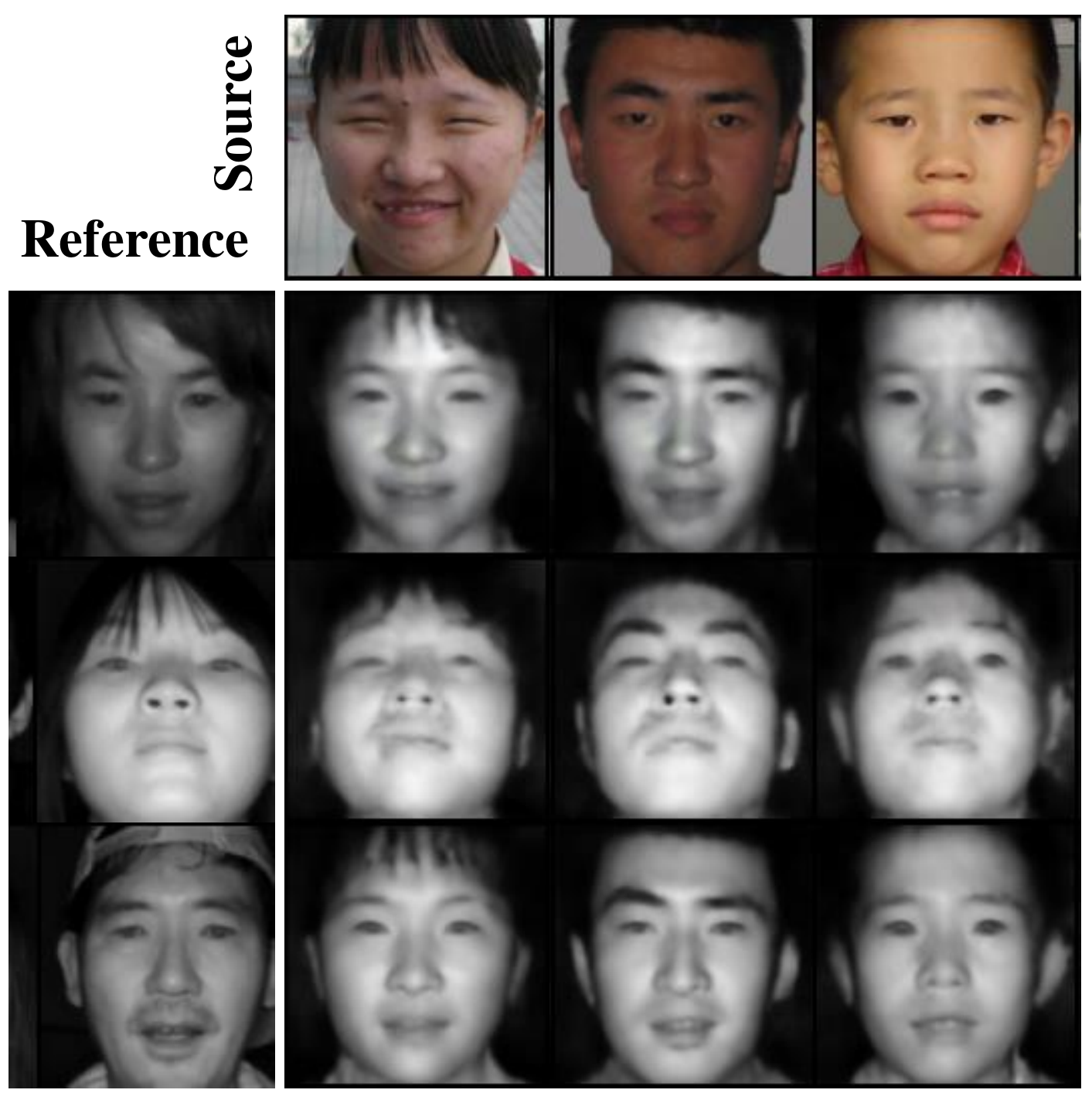}
\caption{The visual results of  cross-domain synthesis. In contrast to Fig. \ref{Cross}, the source images and reference images are not only derived from different subjects, but also belong to different spectral domains.}
\label{Domain}
\end{figure}

For further analysis, we visualize the results of reconstruction and integration to demonstrate the ability of our method. Fig. \ref{Reconstruct} illustrates that the reconstructed images are high-quality and realistic. Besides, we synthesize images with the integration of identity representations and attribute representations. As shown in Fig. \ref{Cross}, the synthesized results maintain the identities of source images and have similar facial attributes with reference images. In the meantime, we explore cross-domain synthesis that source and reference images are derived from different spectral domains. Fig. \ref{Domain} reveals that FSIAD is also applicable to cross-domain synthesis and has outstanding abilities in feature disentanglement and face generation. Note that the source images are different from reference images in both spectral domains and subjects, which raises the difficulty in face synthesis.

% Reconstruction

% grid of results

% Reconstruction

\subsection{Quantitative Analyses}
Quantitative analyses are carried out to evaluate the performance of our proposed method with comparison to the state-of-the-art methods. In this section, quantitative analyses are divided into three parts, including quantitative evaluations on image synthesis, heterogeneous face recognition experiments and ablation studies. 

\subsubsection{Quantitative evaluations on image synthesis}
We conduct quantitative evaluations to compare the effectiveness and efficiency of MUNIT \cite{Huang2018MUNIT}, FaceShifter \cite{Li2020} and our proposed FSIAD. Three metrics are employed for evaluations: Fr\'{e}chet Inception Distance (FID) \cite{Heusel2017}, Structural Similarity Index (SSIM) \cite{Wang2004} and inference speed. FID is a metric to measure the distance between the distribution of real images and that of synthetic images. The lower FID reveals that the model achieves better performance to synthesize high-fidelity images like real images. In the experiments, 10,000 pairs of NIR-VIS images are randomly sampled from the CASIA NIR-VIS 2.0 dataset. Each pair contains source images $\{I_N, I_V\}$ and reference images $\{X_N, X_V\}$. Then 10,000 pairs of NIR-VIS synthetic images $\{\hat{X}_N, \hat{X}_V\}$ are generated by the aforementioned methods. Furthermore, we utilize a pre-trained Inception V3 network \cite{Szegedy2016} to map the real images and the synthetic images into 192-dimensional feature vectors for computing FID \cite{Salimans2016}. As illustrated in TABLE \ref{FID}, our proposed method achieves the best FID values compared with MUNIT and FaceShifter. In other words, our method is preferable to generate realistic and high-quality images for face augmentation.

% Table of FID
\begin{table}[ht]
\caption{The Fr\'{e}chet Inception Distances between distributions of real images on the CASIA NIR-VIS 2.0 dataset and the synthetic images generated by different methods.}
\centering
\begin{tabular}{lcc}
\toprule
\multirow{2}{*}{Method} & \multicolumn{2}{c}{FID$\downarrow$} \\
\cmidrule(r){2-3}
& NIR & VIS \\
\midrule
MUNIT\cite{Huang2018MUNIT} & 17.4735 & 17.5492 \\

FaceShifter\cite{Li2020} & 11.5227 & 9.0366 \\

FSIAD & \textbf{3.2995} & \textbf{2.9839} \\
\bottomrule
\end{tabular}
\label{FID}
\end{table}

Apart from FID, we quantitatively evaluate the performance of feature disentanglement. To assess the similarity of facial attributes between references and synthetic images, we employ the SSIM to measure similarity in structural information between images. Structural information is defined as the attributes of objects in the scene, which is highly adapted for human visual perception \cite{Wang2004}. We measure the SSIM score between references and synthetic images. The higher SSIM score suggests that facial attributes of synthetic images are more similar to those of references. As depicted in TABLE \ref{SSIM}, FSIAD performs better than state-of-the-art methods and achieves the best ability in feature disentanglement. 

% Table of SSIM
\begin{table}[ht]
\caption{The quantitative results of attribute similarity on the CASIA NIR-VIS 2.0 dataset. SSIM denotes the Structural Similarity index.}
\centering
\begin{tabular}{lcc}
\toprule
\multirow{2}{*}{Method} & \multicolumn{2}{c}{SSIM$\uparrow$} \\
\cmidrule(r){2-3}
& NIR & VIS \\
\midrule
MUNIT\cite{Huang2018MUNIT} & 0.2572 & 0.1310 \\

FaceShifter\cite{Li2020} & 0.2294 & 0.1623 \\

FSIAD & \textbf{0.4139} & \textbf{0.3617} \\
\bottomrule
\end{tabular}
\label{SSIM}
\end{table}

To further assess the efficiency of the aforementioned methods, we measure the time consumed by synthesizing 12,480 pairs of images with these methods. Inference speed is measured by the frames per second (FPS) score that equals a ratio of the number of synthesized images to the time spent on synthesis. As shown in TABLE \ref{Speed}, the proposed FSIAD is the most efficient method and particularly outperforms FaceShifter by 11 times. 

% Table of Inference Speed
\begin{table}[ht]
\caption{The inference speeds of different methods. The number of synthesized images is 12,480.}
\centering
\begin{tabular}{lcc}
\toprule
Method & Time(s)$\downarrow$ & Speed(FPS)$\uparrow$\\

\midrule
MUNIT\cite{Huang2018MUNIT} & 65.4009 & 190.823 \\

FaceShifter\cite{Li2020} & 415.9640 & 30.002 \\

FSIAD & \textbf{36.7743} & \textbf{339.367} \\
\bottomrule
\end{tabular}
\label{Speed}
\end{table}

\subsubsection{Heterogeneous face recognition experiments}
In order to validate the effectiveness of FSIAD on heterogeneous face recognition, we conduct extensive experiments on five HFR databases, including CASIA NIR-VIS 2.0 \cite{Li2013CASIA}, BUAA-VisNir \cite{Huang2012BUAA}, Oulu-CASIA NIR-VIS \cite{Chen2009Oulu}, Tufts Face \cite{Panetta2020TUFTS} and LAMP-HQ \cite{Yu2021LAMP}. Our method is compared with the state-of-the-art methods TRIVET \cite{Liu2016}, IDR \cite{He2017}, ADFL \cite{Song2018}, W-CNN \cite{He2019}, PACH \cite{Duan2020}, DVR \cite{Wu2019}, DVG \cite{DVG}, HFIDR \cite{Liu2021NNLS}, and OMDRA \cite{OMDRA}. Since our method utilizes LightCNN \cite{Wu2018LightCNN} as HFR model, the LightCNN only trained with original HFR dataset is selected as a baseline method.
Interestingly, we also train LightCNN \cite{Wu2018LightCNN} with the synthetic face images generated by MUNIT \cite{Huang2018MUNIT} and FaceShifter \cite{Li2020} to explore the effectiveness of these two methods. We denote LightCNN models trained with synthetic faces generated by these two methods as LightCNN+MUNIT and LightCNN+FS, respectively. 
The experiments are conducted by following the protocols as described in Section \ref{sec:database}. The specific experimental analyses are provided as follows.

\begin{figure*}[ht]
\vspace{-10pt}
  \subfigure[CASIA-NIR-VIS 2.0]{
  \begin{minipage}[t]{0.33\textwidth}
    \centering
    \includegraphics[width=\textwidth]{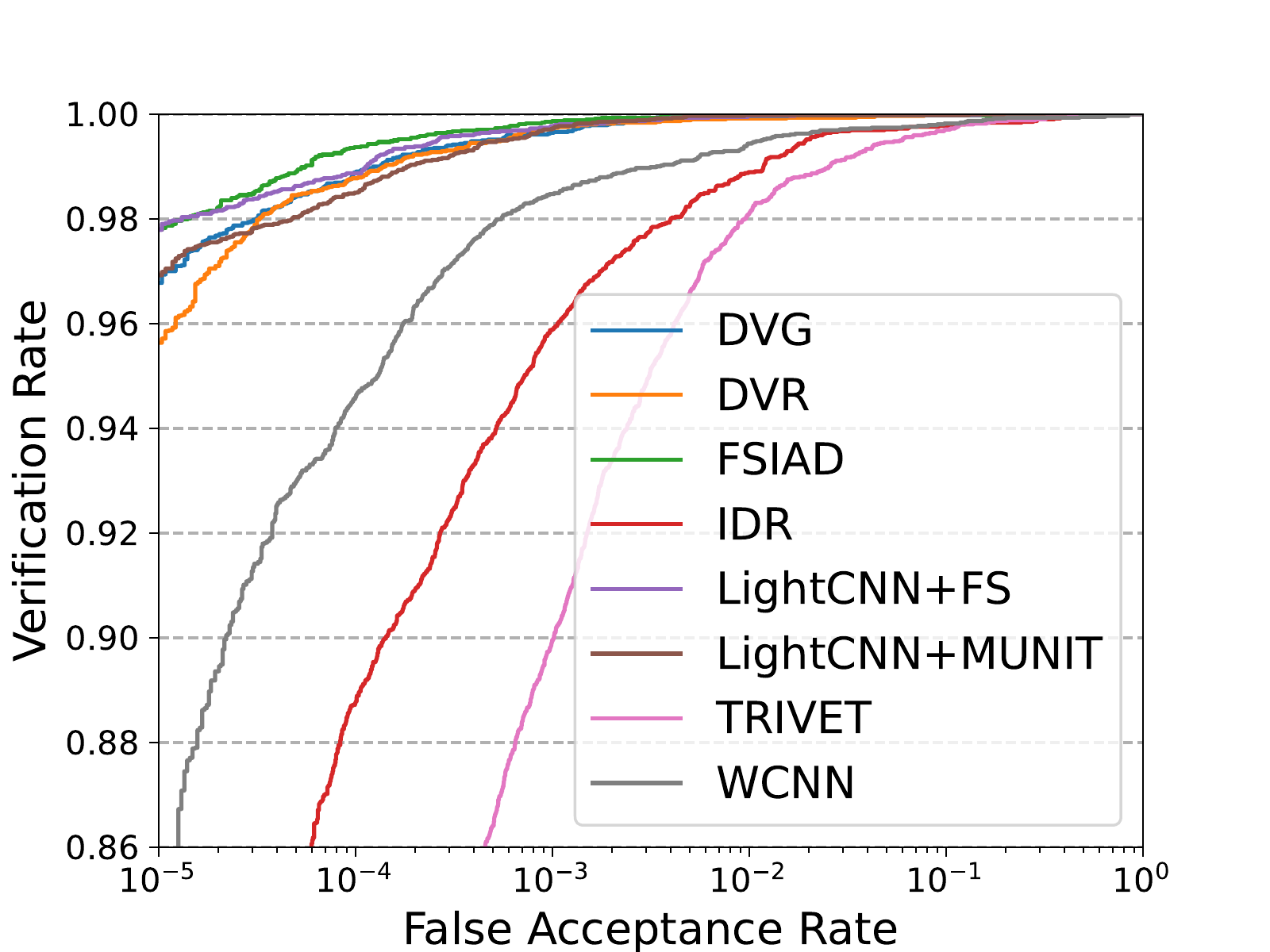}
  \end{minipage}
  \label{fig:roc_casia}
  }
%   \hfill
  \subfigure[BUAA-VisNir]{
  \begin{minipage}[t]{0.33\textwidth}
    \centering
    \includegraphics[width=\textwidth]{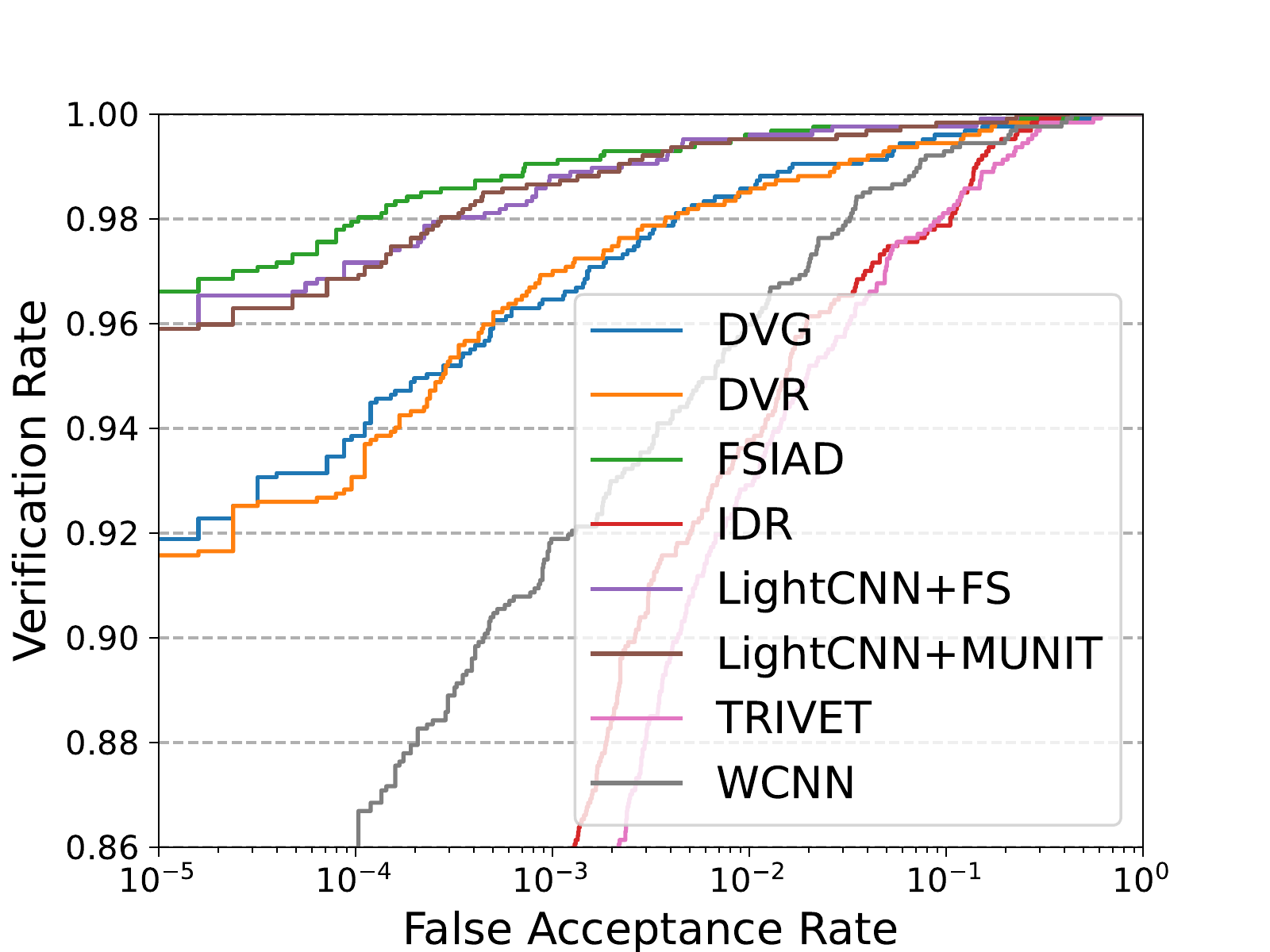}
  \end{minipage}
  \label{fig:roc_buaa}
  }
%   \hfill
  \subfigure[Oulu-CASIA NIR-VIS]{
  \begin{minipage}[t]{0.33\textwidth}
    \centering
    \includegraphics[width=\textwidth]{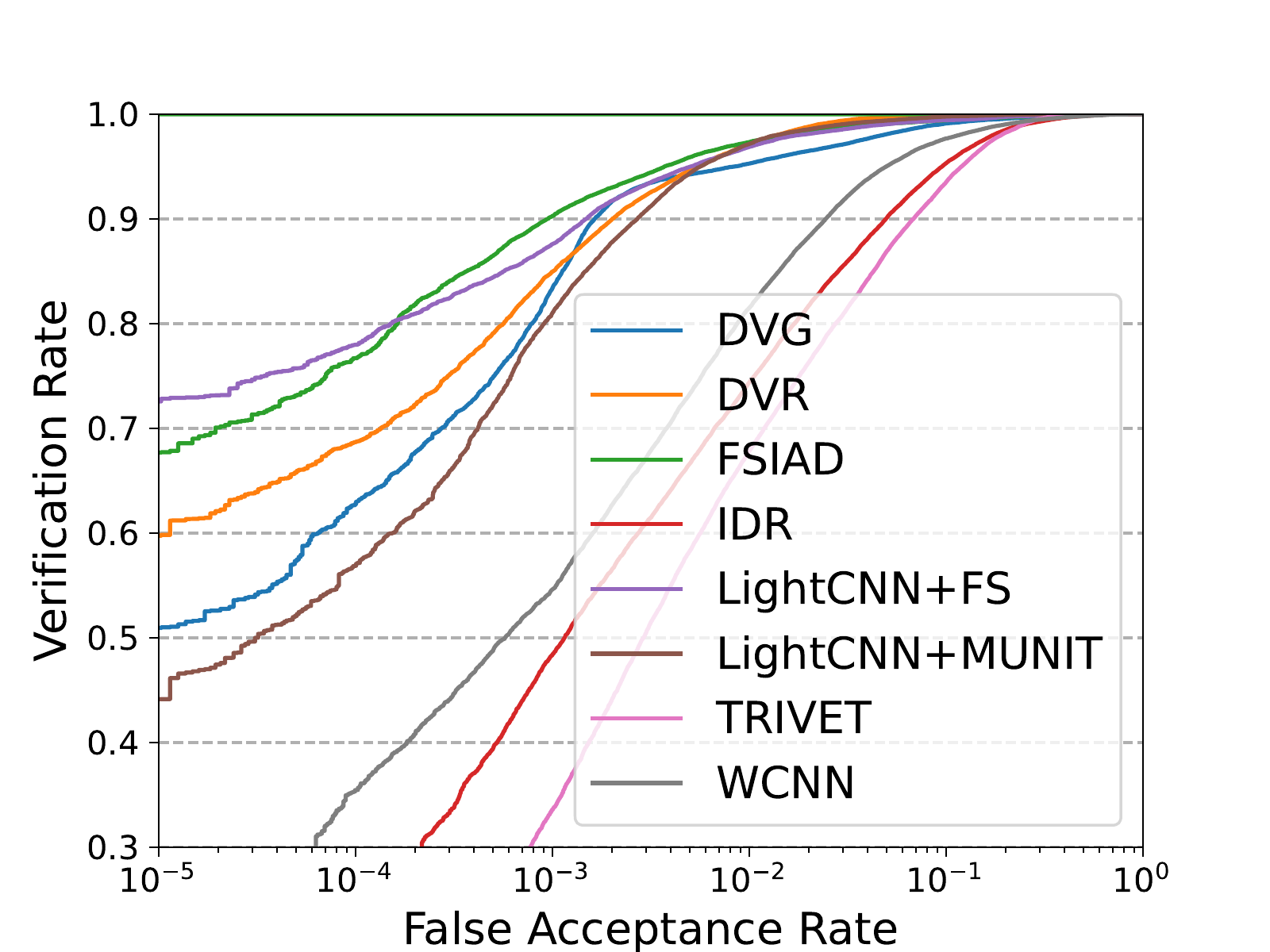}
  \end{minipage}
  \label{fig:roc_oulu}
  }
  \vspace{-10pt}
  \caption{The ROC curves of different methods on the CASIA-NIR-VIS 2.0, BUAA-VisNir, and Oulu-CASIA NIR-VIS datasets.}
  \label{fig:roc}
\vspace{-10pt}
\end{figure*}

\textbf{CASIA NIR-VIS 2.0 database.} We conduct 10-fold cross validation on the CASIA NIR-VIS 2.0 database. The Rank-1 accuracy, Verification Rate(VR)@False Accept Rate(FAR)=0.1\% and VR@FAR=0.01\% are employed for testing. TABLE \ref{CASIA} and Fig. \ref{fig:roc_casia} show that most methods gain satisfactory performance of HFR and achieve Rank-1 accuracy over 90\%. Although HFIDR(LightCNN-9) \cite{Liu2021NNLS} has the lowest Rank-1 accuracy, it can dramatically improve by simply replacing the backbone from LightCNN-9 to LightCNN-29. Compared with TRIVET \cite{Liu2016}, IDR \cite{He2017}, LightCNN \cite{Wu2018LightCNN} and ADFL \cite{Song2018}, advanced methods including W-CNN \cite{He2019}, DVR \cite{Wu2019}, HFIDR \cite{Liu2021NNLS}, OMDRA \cite{OMDRA}, PACH \cite{Duan2020}, DVG \cite{DVG} effectively address the over-fitting problem and obtain better performance. Our proposed FSIAD outperforms the state-of-the-art methods and demonstrates its superiority.
FSIAD surpasses PACH by 1.6\% of VR@FAR=0.1\%, which reveals unconditional face generation provides sufficient heterogeneous faces and fundamentally deals with a key obstacle of HFR. Furthermore, FSIAD outperforms DVG \cite{DVG} by 0.4\% in terms of VR@FAR=0.01\%. It shows that enriching the diversity in facial attributes is conducive to improving HFR. 
We also find that synthetic images generated by MUNIT \cite{Huang2018MUNIT} and FaceShifter \cite{Li2020} facilitate the performance of LightCNN and exceed state-of-the-art methods OMDRA \cite{OMDRA} and HFIDR \cite{Liu2021NNLS}. However, the results of LightCNN+MUNIT and LightCNN+FS fail to surpass DVG due to the poor capabilities of face generation and feature disentanglement. 
Moreover, FSIAD has the smallest standard deviation in the performance of HFR among state-of-the-art methods. Experimental results substantiate that our method effectively tackles the challenges of HFR and facilitates the performance of HFR.

% Table of CASIA 
\begin{table}[ht]
\caption{The 10-fold experimental results of recognition on the CASIA NIR-VIS 2.0 database.}
\centering
\resizebox{\linewidth}{!}{
\begin{tabular}{lccc}
\toprule
Method & \tabincell{c}{Rank-1\\(\%)} & \tabincell{c}{VR@FAR\\=0.1\%(\%)} & \tabincell{c}{VR@FAR\\=0.01\%(\%)} \\

\midrule
TRIVET\cite{Liu2016} & 95.7$\pm$0.5 & 91.0$\pm$1.3 & 74.5$\pm$0.7 \\

IDR\cite{He2017} & 97.3$\pm$0.4 & 95.7$\pm$0.7 & - \\

ADFL\cite{Song2018} & 98.2$\pm$0.3 & 97.2$\pm$0.5 & - \\

W-CNN\cite{He2019} & 98.7$\pm$0.3 & 98.4$\pm$0.4 & 94.3$\pm$0.4 \\

PACH\cite{Duan2020} & 98.9$\pm$0.2 & 98.3$\pm$0.2 & - \\

DVR\cite{Wu2019} & 99.7$\pm$0.1 & 99.6$\pm$0.3 & 98.6$\pm$0.3 \\

DVG\cite{DVG} & 99.8$\pm$0.1 & 99.8$\pm$0.1 & 98.8$\pm$0.2 \\

HFIDR(LightCNN-9)\cite{Liu2021NNLS} & 87.5$\pm$0.0 & - & - \\

HFIDR(LightCNN-29)\cite{Liu2021NNLS} & 98.6$\pm$0.0 & - & - \\

OMDRA\cite{OMDRA} & 99.6$\pm$0.1 & 99.4$\pm$0.2 & - \\

LightCNN\cite{Wu2018LightCNN} & 96.7$\pm$0.2 & 94.8$\pm$0.4 & 88.5$\pm$0.2 \\

LightCNN+MUNIT\cite{Huang2018MUNIT} & 99.6$\pm$0.0 & 99.5$\pm$0.0 & 98.1$\pm$0.1 \\

LightCNN+FS\cite{Li2020} & 99.7$\pm$0.0 & 99.6$\pm$0.0 & 98.1$\pm$0.2 \\

FSIAD & \textbf{99.9$\pm$0.0} & \textbf{99.9$\pm$0.0} & \textbf{99.2$\pm$0.1} \\
\bottomrule
\end{tabular}
}
\label{CASIA}
\end{table}

\textbf{BUAA-VisNir database.} The Rank-1 accuracy, VR@FAR=1\%, and VR@FAR=0.1\% are used for testing on the BUAA-VisNir database. As illustrated in TABLE \ref{BUAA} and Fig. \ref{fig:roc_buaa}, we observe the results of most state-of-the-art methods including TRIVET \cite{Liu2016}, IDR \cite{He2017}, ADFL \cite{Song2018}, W-CNN \cite{He2019}, PACH \cite{Duan2020}, DVR \cite{Wu2019}, and DVG \cite{DVG} are lower than 99\% in terms of VR@FAR=1\%. The possible reason is that BUAA-VisNir is a small-scale HFR dataset and contains images of limited subjects. These methods except DVG suffer from the over-fitting problem and get unsatisfactory performance. Although DVG generates sufficient heterogeneous faces as auxiliary training data, it does not take various facial appearances into consideration and leads to degradation due to variations in the pose and expression of faces on this dataset. LightCNN+MUNIT \cite{Huang2018MUNIT}, LightCNN+FS \cite{Li2020} and the proposed FSIAD enrich the diversity of facial attributes in generated images and achieve outstanding performance, whose results are higher than 99.5\% in terms of VR@FAR=1\%. OMDRA \cite{OMDRA} learns residual-independent identity representations and obtains the best results. The results of FSIAD are close to those of OMDRA. FSIAD reaches 99.7\% in terms of VR@FAR=1\%, which is second to OMDRA (99.9\%).

% Table of BUAA 
\begin{table}[ht]
\caption{The results of recognition on the BUAA-VisNir database.}
\centering
\resizebox{\linewidth}{!}{
\begin{tabular}{lccc}
\toprule
Method & \tabincell{c}{Rank-1\\(\%)} & \tabincell{c}{VR@FAR\\=1\%(\%)} & \tabincell{c}{VR@FAR\\=0.1\%(\%)} \\

\midrule
TRIVET\cite{Liu2016} & 93.9 & 93.0 & 80.9 \\

IDR\cite{He2017} & 94.3 & 93.4 & 84.7 \\

ADFL\cite{Song2018} & 95.2 & 95.3 & 88.0 \\

W-CNN\cite{He2019} & 97.4 & 96.0 & 91.9 \\

PACH\cite{Duan2020} & 98.6 & 98.0 & 93.5 \\

DVR\cite{Wu2019} & 99.2 & 98.5 & 96.9 \\

DVG\cite{DVG} & 99.3 & 98.5 & 97.3 \\

OMDRA\cite{OMDRA} & \textbf{100.0} & \textbf{99.9} & \textbf{99.7} \\

LightCNN\cite{Wu2018LightCNN} & 96.5 & 95.4 & 86.7 \\

LightCNN+MUNIT\cite{Huang2018MUNIT} & 99.4 & 99.5 & 98.7 \\

LightCNN+FS\cite{Li2020} & 99.5 & 99.5 & 98.8 \\

FSIAD & 99.8 & 99.7 & 99.1 \\
\bottomrule
\end{tabular}
}
\label{BUAA}
\end{table}

\textbf{Oulu-CASIA NIR-VIS database.} 
The Rank-1 accuracy, VR@FAR=1\%, and VR@FAR=0.1\% are used for testing. TABLE \ref{Oulu} and Fig. \ref{fig:roc_oulu} report that all methods confront degradation of performance on the Oulu-CASIA NIR-VIS dataset compared with CASIA NIR-VIS 2.0 and BUA-VisNir datasets. Since this dataset contains more types of expressions and illuminations but has fewer subjects than those of the former two datasets, it makes HFR more difficult. The results of TRIVET \cite{Liu2016}, IDR \cite{He2017}, ADFL \cite{Song2018}, W-CNN \cite{He2019} are lower than 80\% in terms of VR@FAR=0.1\%. These methods are weak in generalization for HFR and fail to solve the difficulty of insufficient subjects on this dataset.
Besides, DVR \cite{Wu2019}, PACH \cite{Duan2020}, DVG \cite{DVG} have not considered feature disentanglement and can not adapt to variations in facial attributes for better performance. HFIDF \cite{Liu2021NNLS}, OMDRA \cite{OMDRA}, LightCNN+MUNIT \cite{Huang2018MUNIT}, LightCNN+FS \cite{Li2020} and FSIAD achieve prominent results and gain 100\% in terms of Rank-1 accuracy. 
It is notable that LightCNN+MUNIT and LightCNN+FS improve LightCNN \cite{Wu2018LightCNN} from 65.1 to 81.0\% and 87.6\% in terms of VR@FAR=0.1\%, respectively. This significant improvement indicates that the diverse facial attributes of augmented heterogeneous faces exert a positive influence on HFR. OMDRA achieves the best performance among other methods. FSIAD obtains comparable results to those of OMDRA, where the difference of VR@FAR=0.1\% is 0.2\%.

% Table of Oulu-CASIA 
\begin{table}[ht]
\caption{The results of recognition on the Oulu-CASIA NIR-VIS database.}
\centering
\resizebox{\linewidth}{!}{
\begin{tabular}{lccc}
\toprule
Method & \tabincell{c}{Rank-1\\(\%)} & \tabincell{c}{VR@FAR\\=1\%(\%)} & \tabincell{c}{VR@FAR\\=0.1\%(\%)} \\

\midrule
TRIVET\cite{Liu2016} & 92.2 & 67.9 & 33.6 \\

IDR\cite{He2017} & 94.3 & 73.4 & 46.2 \\

ADFL\cite{Song2018} & 95.5 & 83.0 & 60.7 \\

W-CNN\cite{He2019} & 98.0 & 81.5 & 54.6 \\

DVR\cite{Wu2019} & 100.0 & 97.2 & 84.9 \\

PACH\cite{Duan2020} & 100.0 & 97.9 & 88.2 \\

DVG\cite{DVG} & 100.0 & 97.5 & 90.6 \\

HFIDF(LightCNN-9)\cite{Liu2021NNLS} & 100.0 & - & - \\

HFIDF(LightCNN-29)\cite{Liu2021NNLS} & 100.0 & - & - \\

OMDRA\cite{OMDRA} & 100.0 & \textbf{98.5} & \textbf{92.2} \\

LightCNN\cite{Wu2018LightCNN} & 96.7 & 92.4 & 65.1 \\

LightCNN+MUNIT\cite{Huang2018MUNIT} & 100.0 & 97.2 & 81.0 \\

LightCNN+FS\cite{Li2020} & 100.0 & 97.0 & 87.6 \\

FSIAD & \textbf{100.0} & 98.1 & 92.0 \\
\bottomrule
\end{tabular}
}
\label{Oulu}
\end{table}

\textbf{Tufts Face database.} We compare our method with LightCNN and DVG on the Tufts Face database. TABLE \ref{Tufts} demonstrates that it is challenging to match faces between thermal and VIS domains due to the lack of facial textures and geometries and the large cross-domain discrepancy. Owing to the small-scale training data, LightCNN \cite{Wu2018LightCNN} gets poor results on the Tufts Face dataset and reaches 15.3\% in terms of Rank-1 accuracy. Benefited from synthetic faces, DVG \cite{DVG} increases the Rank-1 accuracy to 51.6\%. It is noted that there are variations in expressions, angles, and accessories such as sunglasses. FSIAD further improves the performance of HFR and reaches 57.1\% in terms of Rank-1 accuracy. The results reveal that augmentation of facial attributes is conducive to Thermal-VIS face recognition. 

% Table of Tufts Face 
\begin{table}[htbp]
\caption{The results of recognition on the Tufts Face database.}
\centering
\begin{tabular}{lcc}
\toprule
Method & Rank-1(\%) & VR@FAR=1\%(\%) \\

\midrule
LightCNN\cite{Wu2018LightCNN} & 15.3 & 6.1 \\

DVG\cite{DVG} & 51.6 & 31.6 \\

FSIAD & \textbf{57.1} & \textbf{35.7} \\
\bottomrule
\end{tabular}
\label{Tufts}
\end{table}

\begin{figure}[ht]
\centering
\includegraphics[width=\linewidth]{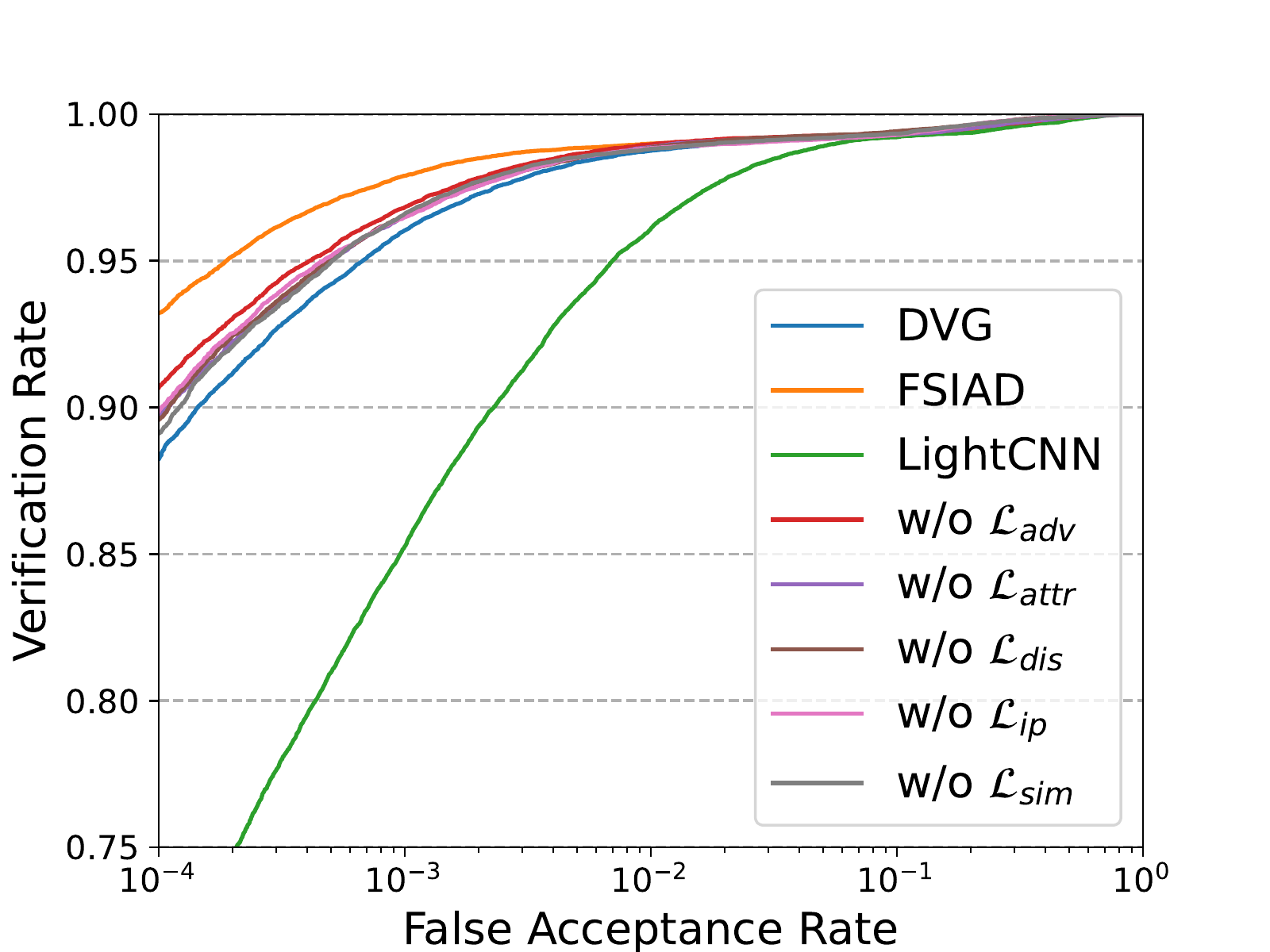}
\caption{The ROC curves of LightCNN, DVG, the proposed FSIAD and its five variants on the LAMP-HQ database.}
\label{ROC}
\end{figure}

\textbf{LAMP-HQ.} We conduct 1-fold and 10-fold experiments on the LAMP-HQ database to validate the effectiveness of our method. Four metrics are used for evaluations, including Rank-1 accuracy, VR@FAR=1\%, VR@FAR=0.1\% and VR@FAR=0.01\%. 
The results of the 1-fold experiment are reported in TABLE \ref{LAMP-HQ1} and Fig. \ref{ROC}.
Conditional synthesis based methods ADFL \cite{Song2018} and PACH \cite{Duan2020} have poor generalization abilities to deal with various facial attributes and get results close to those of baseline LightCNN \cite{Wu2018LightCNN}. DVG \cite{DVG} takes the advantage of unconditional face generation to synthesize sufficient heterogeneous faces as training data. Unfortunately, since DVG is not able to generate faces with diverse facial attributes, large variations in facial attributes hinder the enhancement of its performance. DVG exceeds LightCNN by 2.1\% and 2.7\% in terms of Rank-1 accuracy and VR@FAR=1\%, respectively. FSIAD has made significant progress and outperforms all the state-of-the-art methods. Especially, FSIAD improves Rank-1 accuracy and VR@FAR=0.01\% from 98.3\% to 98.8\% and from 88.2\% to 93.0\%, respectively.
These results prove that the diversity in facial attributes of training data has a critical impact on the improvement of HFR. 
 
% Table of LAMP-HQ 1-fold
\begin{table}[ht]
\caption{The 1-fold experimental results on the LAMP-HQ database. Results of ADFL\cite{Song2018} and PACH\cite{Duan2020} are cited from \cite{Yu2021LAMP}.}
\centering
\resizebox{\linewidth}{!}{
\begin{tabular}{lcccc}
\toprule
Method & \tabincell{c}{Rank-1\\(\%)} & \tabincell{c}{VR@FAR\\=1\%(\%)} & \tabincell{c}{VR@FAR\\=0.1\%(\%)} & \tabincell{c}{VR@FAR\\=0.01\%(\%)} \\

\midrule
LightCNN\cite{Wu2018LightCNN} & 96.2 & 96.1 & 85.3 & 69.3 \\

ADFL\cite{Song2018} & 95.8 & 91.5 & 71.0 & - \\

PACH\cite{Duan2020} & 96.9 & 93.9 & 78.7 & - \\

DVG\cite{DVG} & 98.3 & 98.8 & 96.0 & 88.2 \\

FSIAD & \textbf{98.8} & \textbf{99.1} & \textbf{97.9} & \textbf{93.0} \\
\bottomrule
\end{tabular}
}
\label{LAMP-HQ1}
\end{table}

% ROC

\begin{figure}[ht]
\centering
\includegraphics[width=0.95\linewidth]{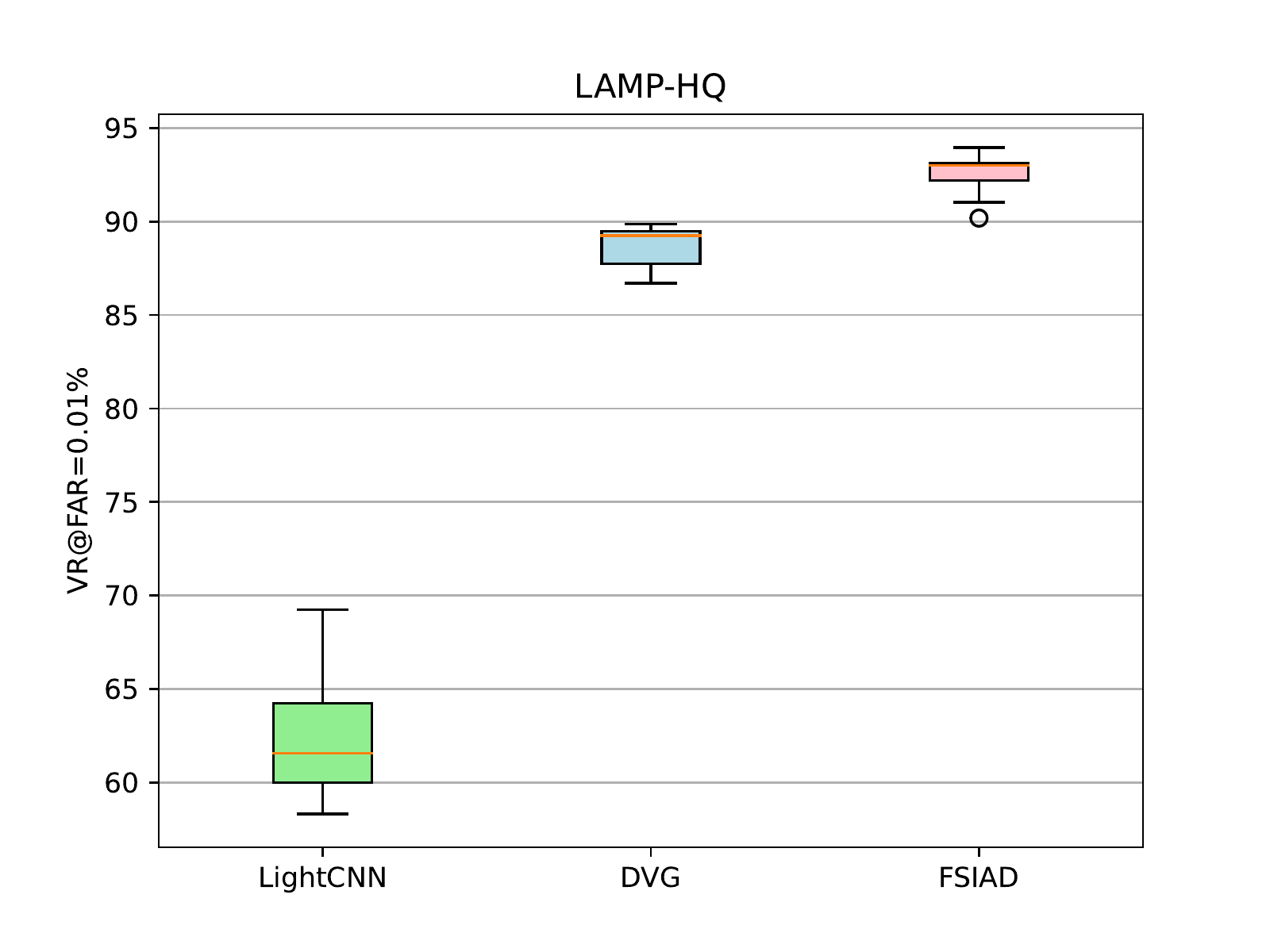}
\caption{The box plots of LightCNN, DVG and the proposed FSIAD on the LAMP-HQ database. }
\label{Box}
\end{figure}

The results of 10-fold experiments are presented in TABLE \ref{LAMP-HQ10} and Fig. \ref{Box}. We observe that conditional synthesis based methods ADFL \cite{Song2018} and PACH \cite{Duan2020} have similar performance with LightCNN \cite{Wu2018LightCNN}. It reveals the limited paired heterogeneous faces hinder the improvement of HFR. Compared with ADFL and PACH, DVG gains progress and improves VR@FAR=1\% from 95.5\% to 99.0\%. The results of DVG suggest that increasing heterogeneous faces by unconditional synthesis boosts the performance of HFR.
FSIAD exceeds state-of-the-art methods, whose Rank-1 accuracy, VR@FAR=1\%, VR@FAR=0.1\%, and VR@FAR=0.01\% are further improved by 0.4\%, 0.2\%, 0.9, and 4.0\%, respectively. Moreover, Fig. \ref{Box} illustrates FSIAD also has stable performance and achieves the minimal standard deviation of results. Experimental results indicate it is fundamental to enrich the attribute diversity of synthetic heterogeneous faces to facilitate HFR. FSIAD produces sufficient heterogeneous faces with abundant facial attributes, which improve the generalization ability of the HFR network and substantially alleviate the degradation of performance caused by large variations in attributes. 

% Table of LAMP-HQ 10-fold
\begin{table}[ht]
\caption{The 10-fold experimental results on the LAMP-HQ database. Results of ADFL\cite{Song2018} and PACH\cite{Duan2020} are cited from \cite{Yu2021LAMP}.}
\centering
\resizebox{\linewidth}{!}{
\begin{tabular}{lcccc}
\toprule
Method & \tabincell{c}{Rank-1\\(\%)} & \tabincell{c}{VR@FAR\\=1\%(\%)} & \tabincell{c}{VR@FAR\\=0.1\%(\%)} & \tabincell{c}{VR@FAR\\=0.01\%(\%)} \\

\midrule
LightCNN\cite{Wu2018LightCNN} & 95.8$\pm$0.1 & 95.5$\pm$0.3  & 82.4$\pm$2.3 & 62.5$\pm$10.4\\

ADFL\cite{Song2018} & 95.1$\pm$0.5 & 92.1$\pm$0.9 & 73.3$\pm$2.2 & - \\

PACH\cite{Duan2020} & 95.4$\pm$0.5 & 93.1$\pm$0.4 & 75.3$\pm$1.7 & - \\

DVG\cite{DVG} & 98.3$\pm$0.1  & 99.0$\pm$0.1 & 96.4$\pm$0.2 & 88.6$\pm$1.5 \\

FSIAD & \textbf{98.7$\pm$0.1} & \textbf{99.2$\pm$0.1} & \textbf{97.3$\pm$0.2} & \textbf{92.6$\pm$1.2} \\
\bottomrule
\end{tabular}
}
\label{LAMP-HQ10}
\end{table}

% Box Plot

\subsection{Ablation Studies}
To investigate the effectiveness of the proposed loss functions in FSIAD, we conduct ablation studies on the LAMP-HQ database.
Since we follow the experimental settings of \cite{Liu2021NNLS,DVG,Wu2019,Duan2020,He2019} to adopt LightCNN \cite{Wu2018LightCNN} as the HFR network, we initialize its parameters with a model pre-trained on the MS-Celeb-1M dataset \cite{Guo2016} as a baseline (BL). Then the baseline model is fine-tuned on the LAMP-HQ dataset and denoted as \texttt{LightCNN} in TABLE \ref{Ablation_Table}. For ablation studies, we construct five variants and each variant is trained without a specific loss function.  

\begin{figure}[ht]
\centering
\includegraphics[width=\linewidth]{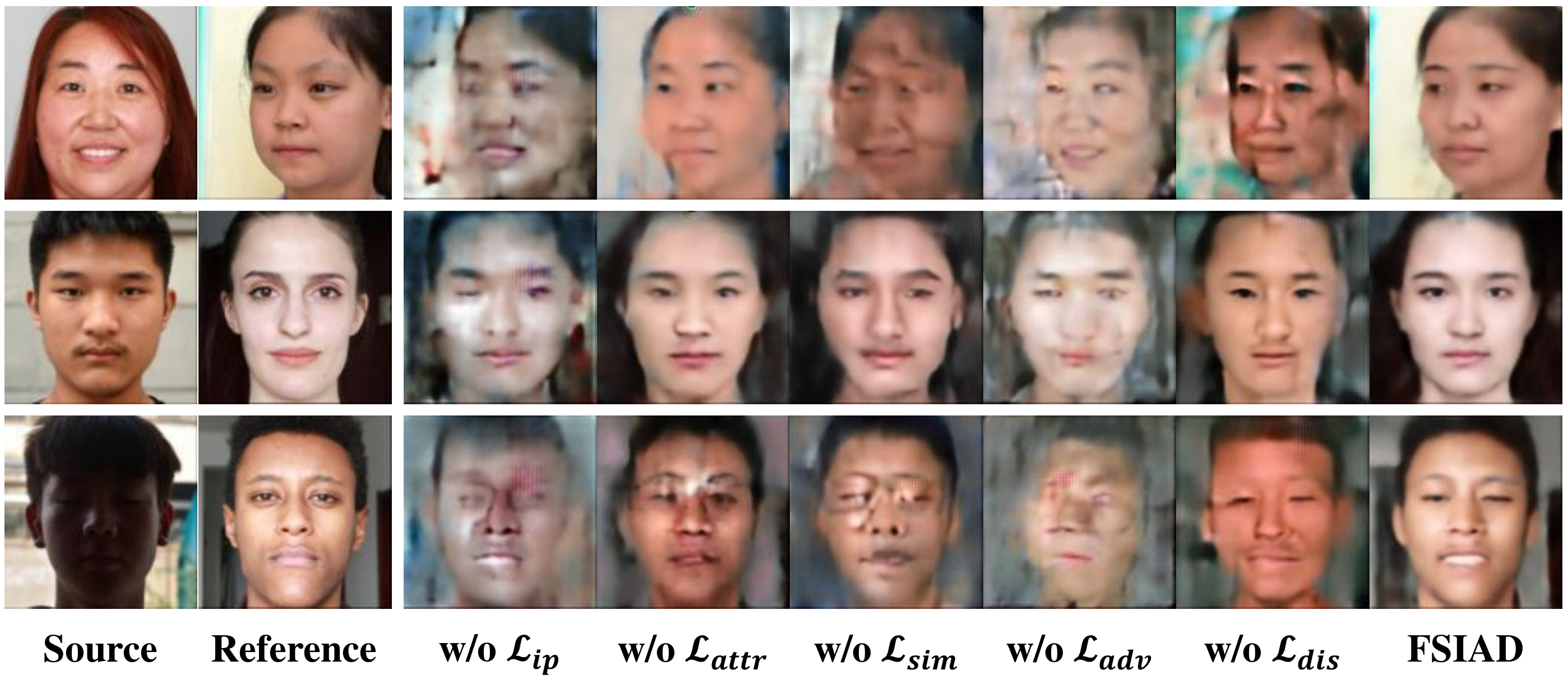}
\caption{The qualitative results of ablation studies on the LAMP-HQ database. The first and second columns are source images and reference images. The images in the rightmost column are synthesized by FSIAD, while the rest columns are synthesized by the ablation versions without loss functions $\mathcal{L}_{ip}$, $\mathcal{L}_{attr}$, $\mathcal{L}_{sim}$, $\mathcal{L}_{adv}$, and $\mathcal{L}_{dis}$, respectively.}
\label{Ablation_Figure}
\end{figure}

Qualitative and quantitative evaluations are conducted. 
We firstly train variants on the LAMP-HQ dataset and then synthesize faces from disentangled representations of identities and attributes. Fig. \ref{Ablation_Figure} illustrates the synthetic faces generated by five variants. Obviously, the faces of FSIAD w/o $\mathcal{L}_{ip}$ are worst and lose most of the information of identity. Without the identity preserving loss $\mathcal{L}_{ip}$, it is difficult for FSIAD to synthesize faces whose identities are consistent with those of source faces. Hence the results of inferior quality indicate that identity preserving loss plays an elementary role in face synthesis. As for facial attributes, we find that the facial attributes are dissimilar with those of reference images when the attribute loss $\mathcal{L}_{attr}$ is removed. For example, we discover that complexions of synthetic faces in the first and second rows are different from those of their corresponding reference faces. Therefore, the attribute loss $\mathcal{L}_{attr}$ can supervise FSIAD to align facial attributes of synthetic faces to those of references. However, only relying on $\mathcal{L}_{attr}$ is not enough for attribute similarity constraint. We observe that the facial textures are blurry and ragged when FSIAD is trained without the structural similarity loss $\mathcal{L}_{sim}$. As we can see in Eq. (\ref{L_sim}), $\mathcal{L}_{sim}$ is composed of a $L_1$-norm loss and a multi-scale structural similarity loss $\mathcal{L}_{MS}$. On the one hand, the $L_1$-norm loss contributes to pixel consistency for face synthesis, and makes global features including color and illumination of synthetic faces similar to those of reference faces. On the other hand, the $\mathcal{L}_{MS}$ loss further regularizes the local features such as eye, mouth, texture, and contour to close with those of references. Consequently, similarities of either attribute representations or structural information can not be dispensed with. The faces generated by FSIAD w/o $\mathcal{L}_{adv}$ are prototypes of synthetic results. But these faces are full of artifacts since the generator lacks the supervision of adversarial discriminator. It indicates that $\mathcal{L}_{adv}$ optimizes FSIAD to refine synthetic images and improve fidelity. Compared with the previous results, FSIAD w/o $\mathcal{L}_{dis}$ produces images of better quality. However, these synthetic faces exist blending boundaries and incoherent textures, especially the first and second images. This phenomenon is caused by the lack of feature disentanglement. 
Without $\mathcal{L}_{dis}$, the attribute representations learned from attribute encoders are mixed up with the identity features of reference faces. It is hard for a generator to produce faces with confused representations and thus result in fuzzy synthetic faces. Compared with variants, FSIAD takes the advantage of our proposed loss functions and successfully produces high-quality faces with the combinations of identity representations and facial attribute representations. The synthetic faces of FSIAD have smooth textures, coherent contours but no noise or artifact. Qualitative ablation studies reveal that all the proposed loss functions are essential for FSIAD to produce high-quality faces and implement heterogeneous face augmentation. 

% Ablation Figure 3.4in

Apart from qualitative analyses, we also conduct quantitative ablation studies on the HFR performance of these variants. As demonstrated in TABLE \ref{Ablation_Table} and Fig. \ref{ROC}, the baseline (BL) of ablation studies performs worst on the LAMP-HQ dataset because it is only pre-trained on the VIS faces. After BL is fine-tuned on LAMP-HQ, it achieves better performance and improves the VR@FAR=1\% from 92.5\% to 96.1\%. To reveal the effectiveness of the synthetic faces, we construct a variant FSIAD w/o $\mathcal{L}_{in}$ by setting the $\gamma$ to 0 in Eq. (\ref{L_HFR}). 
It is equivalent to LightCNN, hence its results are identical to those of LightCNN. 
Owing to the enrichment of facial attributes, all variants of FSIAD outperform LightCNN and DVG. However, FSIAD w/o $\mathcal{L}_{ip}$ neglects the identity preservation and causes inconsistency of identity between source faces and synthetic faces. Hence, it directly damages face recognition and slightly improves VR@FAR=0.1\% to 96.5\%. Due to lacking feature disentanglement of identity and facial attributes, FSIAD w/o $\mathcal{L}_{dis}$ has a flaw in synthesis and causes incoherent textures in the generated faces. It hinders FSIAD from facilitating the performance of HFR and its results are close to those of FSIAD w/o $\mathcal{L}_{ip}$.
Without $\mathcal{L}_{sim}$, FSIAD lacks the structural information of face from the reference image and synthesizes irregular faces. It results in the unsatisfactory performance of HFR. 
Compared with FSIAD w/o $\mathcal{L}_{sim}$, FSIAD w/o $\mathcal{L}_{attr}$ has better ability in face synthesis and improves VR@FAR=0.1\% to 97.0\%.
FSIAD w/o $\mathcal{L}_{adv}$ can basically accomplish the goal of FSIAD that produces faces from identities of source and attributes of reference, and so achieves the best VR@FAR=1\% among the aforementioned variants. But its synthetic faces are of low quality and filled with artifacts and thus lead to the medium performance of HFR. 
FSIAD outperforms the variants and remarkably improves VR@FAR=0.01\% from 90.9\% to 93.0\%. Only when all the proposed loss functions are activated can FSIAD exploit its optimal performance to achieve the best results. In summary, the quantitative and qualitative analyses indicate that all the proposed loss functions in FSIAD are indispensable, and they jointly supervise FSIAD to produce high-quality heterogeneous faces and facilitate the performance of HFR.

% Table of Ablation
\begin{table}[ht]
\caption{The quantitative results of ablation studies on the LAMP-HQ database.}
\centering
\resizebox{\linewidth}{!}{
\begin{tabular}{lcccc}
\toprule
Method & \tabincell{c}{Rank-1\\(\%)} & \tabincell{c}{VR@FAR\\=1\%(\%)} & \tabincell{c}{VR@FAR\\=0.1\%(\%)} & \tabincell{c}{VR@FAR\\=0.01\%(\%)} \\

\midrule
BL & 94.5 & 92.5 & 77.3 & 60.3 \\

LightCNN\cite{Wu2018LightCNN} & 96.2 & 96.1 & 85.3 & 69.3 \\
DVG\cite{DVG} & 98.3 & 98.8 & 96.0 & 88.2 \\

w/o $\mathcal{L}_{in}$ & 96.2 & 96.1 & 85.3 & 69.3 \\
w/o $\mathcal{L}_{ip}$ & 98.5 & 98.8 & 96.5 & 89.9 \\

w/o $\mathcal{L}_{dis}$ & 98.5 & 98.9 & 96.6 & 89.6 \\

w/o $\mathcal{L}_{sim}$ &  98.5 & 98.9 & 96.6 & 89.0 \\

w/o $\mathcal{L}_{attr}$ &  98.6 & 98.9 & 97.0 & 90.9 \\

w/o $\mathcal{L}_{adv}$ &  98.6 & 99.0 & 96.8 & 90.7 \\

FSIAD & \textbf{98.8} & \textbf{99.1} & \textbf{97.9} & \textbf{93.0} \\
\bottomrule
\end{tabular}
}
\label{Ablation_Table}
\end{table}

\section{Conclusions}
\label{Conclusions}
In this paper, we propose a novel method Face Synthesis with Identity-Attribute Disentanglement (FSIAD) to augment heterogeneous face images for facilitating the performance of HFR. FSIAD consists of Identity-Attribute Disentanglement (IAD) and Face Synthesis Module (FSM) components. The IAD component is designed to decouple faces into representations of identities and attributes, where attribute representations are regularized to be uncorrelated with identities. 
Then FSM synthesizes abundant heterogeneous faces from the combinations of disentangled representations of identities and attributes, which augment the insufficient training data of HFR and increase diversity in facial attributes. We train the HFR network with both original HFR data and synthetic heterogeneous faces for improving its performance on face recognition. 
Through disentanglement and synthesis, FSIAD naturally tackles three major challenges of HFR.
Extensive experiments on five HFR databases reveal that our method is superior to previous methods and yields state-of-the-art performance.

% or



\appendices

%\section{}

\section*{Acknowledgment}
The authors sincerely thank the associate editor and the reviewers for their professional comments and suggestions. 

\ifCLASSOPTIONcaptionsoff
  \newpage
\fi



\bibliographystyle{IEEEtran}
\bibliography{TIFS}
\end{document}